\documentclass[lettersize,journal]{IEEEtran}
\usepackage{amsmath,amsfonts}
\usepackage{algorithmic}
\usepackage{algorithm}
\usepackage{array}

\usepackage{textcomp}
\usepackage{stfloats}
\usepackage{url}
\usepackage{verbatim}
\usepackage{graphicx}
\usepackage{subfigure}
\usepackage{multirow}
\usepackage{booktabs}
\usepackage{cite}
\usepackage{xcolor}
\begin{document}

% \title{Robust Principal Component Analysis with Learned Mean and Sample Weights}
\title{Robust Principal Component Analysis via Discriminant Sample Weight Learning}

\author{Yingzhuo Deng, Ke Hu, Bo Li, Yao Zhang
\thanks{Manuscript received XXXX; revised XXXX; accepted
XXXX. Date of publication XXXX; date of current version
XXXX.This work was supported by the Fundamental Research Funds for the Central Universities under grant GK2040260267 and XK2040021004012 \textit{(Corresponding author: Bo Li)}.}

\thanks{Yingzhuo Deng, Ke Hu and Bo Li are with the Key laboratory of Intelligent Technology and Application of Marine Equipment, the Key laboratory of Ship Intelligent System and Technologies, the Key laboratory of Environment Intelligent Perception, the College of Intelligent System Science and Engineering, Harbin Engineering University, Harbin, China (emails: yzdeng@hrbeu.edu.cn; hke@hrbeu.edu.cn; boli@hrbeu.edu.cn).}

\thanks{Yao Zhang is with School of Engineering, University of Southampton, Southampton, UK (e-mail: yao.zhang@soton.ac.uk).}}

\markboth{Journal of \LaTeX\ Class Files,~Vol.~14, No.~8, August~2021}%
{Shell \MakeLowercase{\textit{et al.}}: A Sample Article Using IEEEtran.cls for IEEE Journals}

% \markboth{IEEE Neural Networks and Learning Systems, Vol. xx, No. xx, November 2023}
% {Shell \MakeLowercase{\textit{et al.}}: Bare Demo of IEEEtran.cls for IEEE Journals}

\maketitle


\begin{abstract}
Principal component analysis (PCA) is a classical feature extraction method, but it may be adversely affected by outliers, resulting in inaccurate learning of the projection matrix. 
This paper proposes a robust method to estimate both the data mean and the PCA projection matrix by learning discriminant sample weights from data containing outliers. 
Each sample in the dataset is assigned a weight, and the proposed algorithm iteratively learns the weights, the mean, and the projection matrix, respectively.
Specifically, when the mean and the projection matrix are available, via fine-grained analysis of outliers, a weight for each sample is learned hierarchically so that outliers have small weights while normal samples have large weights. 
With the learned weights available, a weighted optimization problem is solved to estimate both the data mean and the projection matrix. 
Because the learned weights discriminate outliers from normal samples, the adverse influence of outliers is mitigated due to the corresponding small weights. 
Experiments on toy data, UCI dataset, and face dataset demonstrate the effectiveness of the proposed method in estimating the mean and the projection matrix from the data containing outliers.

% \textcolor{red}{
% This research introduces a resilient approach for acquiring knowledge from data that is tainted by outliers. In this method, each entry in the dataset receives a specific weight, and the algorithm put forward in this paper progressively refines these weights, along with determining the dataset's mean and the PCA projection matrix.
% In particular, once the mean and projection matrix are at our disposal, we meticulously scrutinize outliers. A hierarchical learning process assigns weights to individual samples, giving significant weight to outliers and lower weight to regular data points.
% With these learned weights in hand, we tackle an optimization problem that is solved with consideration for the assigned weights. This procedure aids in estimating both the data mean and the projection matrix. As a result of the emphasis placed on outlier identification during the weight learning, the influence of outliers is mitigated due to their correspondingly low weight values.
% The effectiveness of the proposed approach in estimating the mean and projection matrix from data containing outliers is supported by experiments conducted on toy data, the UCI dataset, and a dataset of facial images.}
\end{abstract}


\section{Introduction}
\IEEEPARstart{I}n real-world scenarios, the dimension of the data is high, which leads to high computational expenses during data processing. 
% Besides that, noises and outliers in the data degrade the performance of data processing algorithms or systems. 
To address these challenges, dimensionality reduction emerges as an effective approach. 
Principal component analysis (PCA)\cite{StandardPCA} is a well-established technique for dimensionality reduction, frequently applied in machine learning and signal processing applications\cite{face}\cite{CV}. 
PCA learns a projection matrix that transforms data from a high-dimensional space to a lower-dimensional space while aiming to either preserve the variance of the data or minimize the errors in data reconstruction.
In this way, PCA reduces data dimensionality and information redundancy while retaining most of the essential information, resulting in minimal information loss.

Regrettably, the optimization function utilized in PCA relies on the $L_2$-norm, which has a significant drawback, namely, its tendency to excessively amplify the impact of significant outliers. 
Consequently, these outliers stew the principal components away from their true directions, ultimately compromising the performance of PCA.
% Unfortunately, the objective function of PCA is based on the $L_2$-norm, while one of the shortcoming of the $L_2$-norm is easily exaggerating the effect of large outliers, which is why the outliers deviate the algorithm from the original solution and degrade the performance. 
Recently, there have been numerous researches devoted to improving the robustness of PCA against outliers. 
These methods can be broadly categorized into three main approaches.
The first approach separates the original data into low-rank and sparse components.
The second approach replaces the $L_2$-norm with other norms which are more robust to outliers.
The third approach identifies the outliers through weight learning.

% \textcolor{red}{Regrettably, the optimization function utilized in PCA relies on the $L_2$-norm, which possesses a drawback—namely, its tendency to excessively amplify the impact of significant outliers. Consequently, these outliers can steer the algorithm off course from its intended solution, ultimately compromising its performance.
% In recent times, a multitude of research efforts have been directed towards enhancing the algorithm's resilience in the face of outliers. The strategies for robust PCA can be broadly categorized into three main approaches. The first approach involves retrieving both the low-rank and sparse components from the original dataset. The second approach centers around adopting alternative criteria to restrict the extent of reconstruction errors. Lastly, the third approach introduces a weighted term to effectively discriminate against outliers.
% }

% PCA is a linear algorithm, while it can't solve the nonlinear problem well. Specifically, in Kernel PCA(KPCA)\cite{KPCA}, the original data are mapped to a high-dimensional feature space through kernel trick before using PCA method. 

The first approach assumes that such data are inherently with low-dimensional structures. 
Recently, some researchers aimed to recover both the low-rank and sparse components from the data.
In \cite{RPCA}\cite{CRPCA}, the data is modeled as $\boldsymbol{X}=\boldsymbol{L}+\boldsymbol{S}$ with $\boldsymbol{L}$ and $\boldsymbol{S}$ denote a low-rank matrix and a sparse matrix respectively.
% They introduced the iterative thresholding approach to solve a relaxed convex form of the problem, as $\min\limits_{\boldsymbol{{L},\boldsymbol{S}}}\left\|\boldsymbol{L}\right\|_*+\lambda\left\|\boldsymbol{S}\right\|_1$, where $\left\|\cdot\right\|_*$ represents the nuclear norm and $\left\|\cdot\right\|_1$ represents the $L_1$-norm.
Considering the prevalence of noises in real-world observations, 
stable principal component pursuit (SPCP) \cite{SPCP} builds upon the assumption that data can be decomposed into low-rank and sparse components by adding an additional component that accounts for noise in the principal component subspace like $\boldsymbol{X}=\boldsymbol{L}+\boldsymbol{S}+\boldsymbol{Z}$.
Brahma et al.\cite{RRPCP} proposed a new model that further introduces a noise component in the subspace orthogonal to the principal component subspace.
However, this kind of method assumes that outliers are sparse, in real-world applications, if the outliers are not sparse or follow a specific pattern, the performance can degrade significantly.

In the second approach, the $L_2$-norm is replaced by other norms to improve the robustness to outliers.
% Compared with $L_2$-norm, $L_1$-norm is robust against outliers in the data.
Ke and Kanade proposed $L_1$-PCA\cite{l1PCA}, which replaces the $L_2$-norm with the $L_1$-norm to minimize the reconstruction error. 
% where the objective function to minimize is $\|\boldsymbol{X}-\boldsymbol{PP}^T\boldsymbol{X}\|_1$ with $\boldsymbol{P}$ denoting the projection matrix to optimize.
% From the perspective of projection variance, Kwak proposed PCA-$L_1$ method\cite{PCAL1} focusing on maximizing the projection variance, and the projection matrix is obtained by maximizing $\|\boldsymbol{P}^T\boldsymbol{X}\|_1$.
Kwak et al.\cite{PCAL1} proposed PCA-$L_1$ method by maximizing the variance calculated by $L_1$-norm metric.
Kawk\cite{PCALP} introduced the $L_p$-norm ($0<p\leq 1$) to PCA which is more general than $L_1$-norm. 
% The proposed optimization problem is to maximize $\|\boldsymbol{P}^T\boldsymbol{X}\|_p^p$.
However, the resulting solutions from these methods lack rationality invariance \cite{rotation} and can be difficult to solve.
To address this problem, Ding et al.\cite{R1PCA} proposed $R_1$-PCA which uses $L_{2,1}$-norm of reconstruction errors as metrics. 
This approach retains rotational invariance and demonstrates excellent performance.
% where the optimization problem is formulated as $\min\limits_{\boldsymbol{P}}\sum\limits_{i=1}^n\sqrt{\boldsymbol{x}_i^T\boldsymbol{x}_i-\boldsymbol{x}_i^T\boldsymbol{PP}^T\boldsymbol{x}_i}$. 
Based on $R_1$-PCA, RPCA-OM\cite{PCAOM} is proposed to find the optimal mean of data.
Additionally, $R_1$-norm based methods employ a greedy strategy. 
Nie et al.\cite{nongreedyl21}\cite{nogreedyl212}  proposed a non-greedy algorithm to address the drawbacks of the $R_1$-PCA solution.
Reasoning that $R_1$-norm struggles with heavy outliers\cite{L2PPCA}, Wang et al. 
\cite{L2PPCA,2attentivePCA} extended it to the $L_{2,p}$-norm as a measure of reconstruction error in PCA. 
In \cite{adaptiveLossPCA}, a novel orthogonal $L_{2,0}$-norm is proposed as a constraint to select the most valuable features in PCA. 
However, determining the optimal hyperparameter value for this norm often remains challenging.
 % The objective is to minimize $\|\boldsymbol{X}-\boldsymbol{PP}^T\boldsymbol{X}\|_2^p$.
 % An iterative re-weighted algorithm was proposed to solve the optimization problem. 
Wang et al.\cite{anglePCA} proposed a method called Angle-PCA, which aims to maximize the sum of the ratio of projection variance to reconstruction error.
% with the optimization function as $\max\limits_{\boldsymbol{P}}\sum_{i=1}^n\frac{\|\boldsymbol{P}^T\boldsymbol{x}_i\|_2^1}{\|\boldsymbol{x}_i-\boldsymbol{PP}^T\boldsymbol{x}_i\|_2^1}$.
Angle-PCA improves robustness to outliers, however, it may fail to effectively capture the overall structural features of the data due to its non-linear cost function \cite{PCADI}.

The third approach enhances the robustness to outliers by assigning weights to data points. 
Koren and Carmel\cite{WPCA} introduced Weighted-PCA, which improves the robustness of PCA by assigning small weights to outliers. 
This method utilizes a Laplacian matrix, where the elements represent the similarity between pairs of data points. 
However, it is worth noting that this approach may tend to yield locally optimal solutions.
Zhang et al.\cite{PCAAN} introduced a general framework named Robust Weight Learning with Adaptive Neighbors (RWL-AN), which automatically derives an adaptive weight vector by integrating robustness and sparse neighbor considerations. 
This framework aims to differentiate between correct samples and outliers, providing a more reasonable approach for robustness learning compared to previous works.
Nie et al.\cite{TRPCA} proposed Truncated Robust Principal Component Analysis (T-RPCA) model, which employs an implicitly truncated weighted learning scheme. 
% However, it is important to note that these methods heavily rely on prior knowledge, specifically the number of outliers, which can be challenging to obtain in practical scenarios.
Nie et al. \cite{DRPCA} proposed a method named Discrete Robust PCA (DRPCA), which utilizes self-learning binary weights to identify normal samples iteratively and update principal components accordingly.
Similarly, Wang et al. \cite{pcajrp} extended Angle PCA \cite{anglePCA} by employing discrete weights, assigning a weight of 1 to normal samples and 0 to outliers.
Gao et al.\cite{PCADI} proposes a Robust PCA based on Discriminant Information (RPCA-DI) that enhances the robustness and generalization of PCA against outliers by employing an entropy-regularized sample representation model and a normalized weighting approach in a two-step outlier identification and processing framework.
The aforementioned methods treat normal samples and outliers distinctly. However, they are overly simplistic in their assumptions regarding the distribution of outliers, typically presuming that outliers exhibit global consistency\cite{PCADI}. Although RPCA-DI considers various sources of outliers, its exploration of noise sources remains limited.

Inspired by the distinctions between outliers and normal samples, we propose a \textbf{R}obust \textbf{P}rincipal \textbf{C}omponent \textbf{A}nalysis algorithm via \textbf{D}iscriminant \textbf{S}ample \textbf{W}eight \textbf{L}earning (RPCA-DSWL).
The proposed algorithm leverages learned sample weights to mitigate the adverse influence of outliers in learning the projection matrix and sample mean.
% With the projection matrix and sample mean available, we categorize outliers in three categories. 
% Correspondingly, three sets of weights are learned.
% The final set of weights are calculated by combing these three sets of weights.
% As for the final weights, normal samples are with large weight, while outliers are with small weight. 
% Consequently, when optimizing the projection matrix and sample mean using these final weights, the influence of outliers is non-significant.
The proposed algorithm iteratively learns the sample weights and optimizes the projection matrix and sample mean. The contributions of this research paper are as follows: 

(1) We group outliers into three categories when the PCA projection matrix is available. The three categories are outliers in the principal component subspace, outliers in the subspace complementary to the principal component subspace, and outliers in the union of these two subspaces. 

(2) When the PCA projection matrix is available, we propose an approach to learn three sets of weights corresponding to three categories of outliers. We also propose to merge these three sets of weights into a single set of weights.

(3) We propose an iterative algorithm to learn sample weights, PCA projection matrix, and the data mean sequentially. 

(4) Experiments on the toy data, UCI dataset, and face dataset show that the proposed RPCA-DSWL algorithm is capable of assigning small weights to outliers, and as a consequence, alleviates the adverse effect of outliers on the estimation of the PCA projection matrix and the data mean.

The remainder of the paper is organized as follows: 
Section \uppercase\expandafter{\romannumeral2} reviews the relevant robust PCA algorithms;
Section \uppercase\expandafter{\romannumeral3} derives the proposed algorithm in robustly estimating the PCA projection matrix and the data mean via sample weight learning; 
Section \uppercase\expandafter{\romannumeral4} shows the experimental results using both the toy dataset and real-world datasets; 
Section \uppercase\expandafter{\romannumeral5} gives the conclusion of this paper.

\section{Related Work}
Considering a set of $n$ samples that constitute a data matrix $\boldsymbol{X}=[\boldsymbol{x}_1,\boldsymbol{x}_2,\cdots,\boldsymbol{x}_n] \in\mathbb{R}^{d\times n}$, where $\boldsymbol{x}_i\in \mathbb{R}^{d\times 1}$ denotes the $i$-th sample and $d$ represents the dimensionality of each data sample. 
Assuming that the data samples have been centralized to have a mean of zero. 
The objective of Principal Component Analysis (PCA) is to find a semi-orthogonal projection matrix $\boldsymbol{P}\in\mathbb{R}^{d\times k}$ by minimizing the squared reconstruction errors,  i.e., to solve the following optimization problem:

% \textcolor{red}{Considering a set of $n$ samples that constitute a data matrix $\boldsymbol{X}=[\boldsymbol{x}_1,\boldsymbol{x}_2,\cdots,\boldsymbol{x}_n] \in\mathbb{R}^{d\times n}$, where $\boldsymbol{x}_i\in \mathbb{R}^{d\times 1}$ denotes the $i$th sample and $d$ represents the dimensionality of each data sample. For the sake of simplicity, let's assume that the data samples have been centralized to have a mean of zero. The objective of Principal Component Analysis (PCA) is to identify an orthogonal projection matrix $\boldsymbol{P}\in\mathbb{R}^{d\times m}$ by minimizing the errors in reconstruction. This entails solving the subsequent optimization problem:

\begin{align}
\label{PCAerror}
\min_{\boldsymbol{P}} \sum_{i=1}^{n} \|\boldsymbol{x}_i-\boldsymbol{PP}^T\boldsymbol{x}_i\|_2^2\quad s.t.~ \boldsymbol{P}^T\boldsymbol{P} = \boldsymbol{I}.
\end{align}

% Where $\boldsymbol{I}$ is the identity matrix. Then for any matrix $\boldsymbol{P}$ with $\boldsymbol{P}^T\boldsymbol{P} = \boldsymbol{I}$, we have the Pythagorean theorem in the feature space:

Due to the semi-orthogonality of $\boldsymbol{P}$, there is
\begin{equation}
\sum_{i=1}^{n}\|\boldsymbol{x}_i-\boldsymbol{PP}^T\boldsymbol{x}_i\|_2^2+\sum_{i=1}^{n}\|\boldsymbol{P}^T\boldsymbol{x}_i\|_2^2=\sum_{i=1}^{n}\|\boldsymbol{x}_i\|_2^2.
\end{equation}

Thus, the projection matrix $\boldsymbol{P}$ can be equivalently optimized by maximizing the variance of the projected data, e.g.,
\begin{align}
\label{PCAvar}
\max_{\boldsymbol{P}} \sum_{i=1}^n \|\boldsymbol{P}^T\boldsymbol{x}_i\|_2^2\quad s.t. ~\boldsymbol{P}^T\boldsymbol{P} = \boldsymbol{I}.
\end{align}

% Based on the above analysis, problem (\ref{PCAerror}) and problem (\ref{PCAvar}) are equivalent, which guarantees that the traditional PCA can obtain the global optimal solution by choosing corresponding geometric (minimizing the reconstruction error) or statistical interpretation (maximizing the data variance).

The optimization problem (\ref{PCAvar}) optimizes the projection matrix by maximizing the variance of projected samples. 
In either (\ref{PCAerror}) or (\ref{PCAvar}), the objective function is defined by $L_2$-norm, which is sensitive to outliers, which are samples that deviate from the normal samples significantly. 
To solve this problem, many PCA algorithms were proposed based on other metrics. 
For instance, L1-PCA\cite{l1PCA} and PCA-L1\cite{PCAL1} solve $L1$-norm problems, which are  
\begin{align}
\label{PCAL1}
&\min_{\boldsymbol{P}} \sum_{i=1}^{n} \|\boldsymbol{x}_i-\boldsymbol{PP}^T\boldsymbol{x}_i\|_1\quad s.t.~ \boldsymbol{P}^T\boldsymbol{P} = \boldsymbol{I}, 
\end{align}
and
\begin{align}
\label{PCA-L1}
&\max_{\boldsymbol{P}} \sum_{i=1}^n \|\boldsymbol{P}^T\boldsymbol{x}_i\|_1\quad s.t. ~\boldsymbol{P}^T\boldsymbol{P} = \boldsymbol{I}, 
\end{align}
respectively.

\cite{L2PPCA} proposes to solve the following $L_{2,p}$-norm problem to estimate the PCA projection matrix, i.e.,
\begin{align}
\label{PCALP}
\min_{\boldsymbol{P}} \sum_{i=1}^n \|\boldsymbol{x}_i-\boldsymbol{PP}^T\boldsymbol{x}_i\|_2^p\quad s.t. \boldsymbol{P}^T\boldsymbol{P} = \boldsymbol{I}.
\end{align}

% By changing the form of the norm, the influence of outliers on the solution is mitigated to a certain extent. But when outliers are significant, the robustness of the proposed methods are far from satisfactory.

Moreover, the traditional PCA neglects the estimation of the sample mean, which may also be greatly affected by outliers.
To obtain the estimate of the mean, Nie et al. \cite{PCAOM} proposed an algorithm by minimizing the reconstruction errors with optimal mean, i.e. solving the following problem:
\begin{align}
\label{rpca1}
\min_{\boldsymbol{m},\boldsymbol{P}} \sum_{i=1}^{n} \|(\boldsymbol{I}-\boldsymbol{PP}^T)(\boldsymbol{x}_i-\boldsymbol{m})\|_2\quad s.t.~\boldsymbol{P}^T\boldsymbol{P} = \boldsymbol{I},
\end{align}
where $\boldsymbol{m}\in \mathbb{R}^{d\times 1}$ is the optimal estimate of the sample mean. In this optimization problem, all samples are not centralized. 
\begin{comment}
RPCA-OM uses an iterative re-weighted method to solve the problem (\ref{rpca1}). In each iteration, the problem to be solved is as follow:
\begin{align}
\label{rpca2}
\min_{\boldsymbol{m},\boldsymbol{P}} \sum_{i=1}^n w_i \|(\boldsymbol{I}-\boldsymbol{PP}^T)(\boldsymbol{x}_i-\boldsymbol{m})\|_2^2\quad s.t.~\boldsymbol{P}^T\boldsymbol{P} = \boldsymbol{I}, 
\end{align}
where $w_i=\frac{1}{2\|(\boldsymbol{I}-\boldsymbol{PP}^T)(\boldsymbol{x}_i-\boldsymbol{m})\|_2}$ acts as a weight in the objective function (\ref{rpca2}). 
However, significant outliers still adversely affect the estimate of both mean and projection matrix.
\end{comment}

\section{Robust PCA via Discriminant Sample Weight and Optimal Mean}
% \subsection{Robust Weight Learned from Subspace}

In conventional PCA, each sample's contribution to the objective function in either (\ref{PCAerror}) or (\ref{PCAvar}) is treated as equal. 
However, this approach makes abnormal samples, which either result in significant variance or significant reconstruction errors, exert undue influence over the objective function.
Consequently, the resulting projection matrix becomes skewed by these abnormal samples or outliers.
Furthermore, it is typically assumed that samples have been centralized, namely, the data's mean being zero.
Nonetheless, in reality, the estimate of the mean is also adversely affected by these outliers.
In scenarios involving outliers in the data, it is essential to have a robust algorithm to estimate both the PCA projection matrix and the data mean.

\begin{figure}
    \centering
    \includegraphics[width=3.2in]{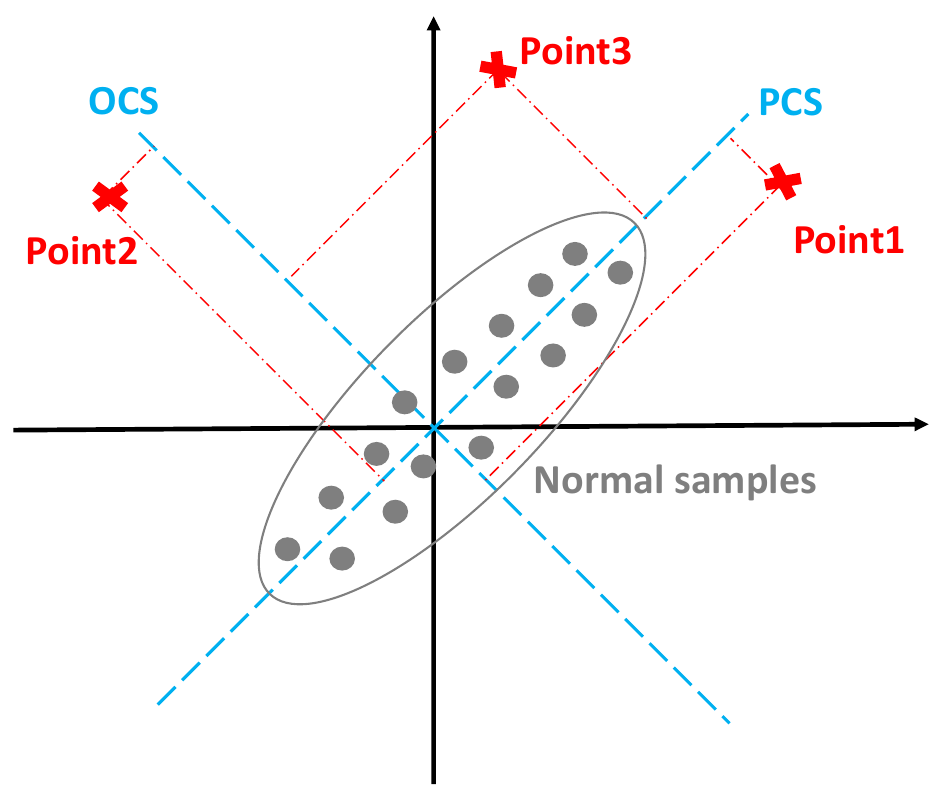}
    \caption{ Illustration of three categories of outliers.  $\textup{Poin1}$, $\textup{Point2}$ and $\textup{Point3}$ represent the first, the second and the third category of outliers, respectively. PCS represents Principal Component Subspace. OCS represents Orthogonal Complement Subspace, which is orthogonal to PCS.}
    \label{fig_1}
\end{figure}

Outliers are data samples whose patterns or feature distributions differ significantly from those of normal samples. 
Concerning principal components, we categorize outliers into three groups.
Fig. \ref{fig_1} illustrates these three categories of outliers.
The first category encompasses outliers lying in the principal component subspace (PCS) or principal subspace, but deviating from normal samples, i.e., having large variances within the subspace. Point1 in Fig. \ref{fig_1} is an example of an outlier belonging to this category.
In the second category, outliers are in the subspace orthogonal to the principal subspace, i.e., the orthogonal complement subspace (OCS). 
These samples exhibit small variances when projected onto the principal subspace but yield large reconstruction errors. 
Point2 in Fig.
\ref{fig_1} is an illustration of this second category of outliers.
The third category of outliers encompasses samples lying in both the principal subspace and its orthogonal complement subspace. 
Although their projection variances and reconstruction errors are intermediate, their distribution patterns deviate from those of normal samples.
Point3 in Fig. \ref{fig_1} belongs to the third category. 
It lies within both the principal subspace and its orthogonal complement, but it is far from the data center.

The distribution of aforementioned outliers differs from normal samples.
The difference could be leveraged to distinguish outliers from normal samples.
In this paper, we propose a new approach to simultaneously estimate the PCA projection matrix, the sample mean, and sample weights.
In the proposed approach, each data sample is assigned a positive weight to distinguish between normal samples and outliers, wherein normal samples are assigned large weights while outliers are assigned small weights. 
% These weights can be utilized to discriminate outliers from normal samples. 
% In the sequel, we will propose an algorithm to robustly learn sample weights and estimate the principal components as well as the optimal mean from the data. 
The algorithm operates iteratively, with each iteration comprising two steps. 
In the first step, concerning each category of the aforementioned outliers, an optimization problem is formulated respectively, assigning a weight to each sample. 
Consequently, each sample is associated with three weights corresponding to three categories of outliers. 
Then, a final weight for each sample is calculated by combining these three individual weights. 
In the second step, using these final weights, another optimization problem is solved to estimate both the PCA projection matrix and the optimal mean. 
These two steps iterate alternatively until convergence is reached.

\subsection{Learning weights for data samples}
As mentioned, the proposed RPCA-DSWL iteratively learns sample weights and the projection matrix as well as the sample mean. 
This part only focuses on learning sample weights, with the PCA projection matrix and data mean available.
In this case, we proposed to assign a weight to each data sample to differentiate outliers from normal data.
For each data sample, three weights are assigned, corresponding to the three categories of outliers outlined above, to one sample. 

The first category of outliers encompasses samples with large variances when projected onto the principal subspace. 
Based on this characteristic, we proposed to solve the following optimization problem to learn a weight for each data sample:

% \textcolor{red}{With the PCA projection matrix in place, we suggest the introduction of weights for data samples as a means to differentiate outliers from regular data.
% For every data sample, a set of three weights is assigned, aligning with the three aforementioned categories of outliers that were outlined earlier.
% The initial outlier category encompasses samples that exhibit substantial variances upon projection onto the principal subspace. Building upon these distinctive attributes, we present the resolution of the subsequent optimization problem in order to acquire a weight for each data sample:}

\begin{equation}
\begin{aligned}
\label{WeightedmeanVAR}
&\max_{a_1,a_2,\cdots,a_n} \frac{1}{n}\sum_{i=1}^{n} a_i \|\boldsymbol{P}^T(\boldsymbol{x}_i-\boldsymbol{m})\|_2^2 + \tau_a\left(-\sum_{i=1}^n a_i \textup{ln}a_i\right)\\
&s.t.\quad\sum_{i=1}^{n} a_i=1,\quad0\leq a_i\leq 1,
\end{aligned}
\end{equation}
where $a_i$ is a weight for the $i$-th sample, and $\tau_a$ is the hyper-parameter to balance the projection variances and the regularizer. 
The regularizer is the entropy of weights, preventing the optimization problem yielding non-trivial solutions.  
The constraints ensure that all weights are both positive and normalized.
Since the optimization problem is a maximization problem, samples with large projection variances are associated with large weights.
In other words, large weights imply large projection variances within the principal subspace, which is the characteristic of the outliers belonging to the first category.

% \textcolor{red}{The regularizer employed is the entropy of the weights, serving to prevent the optimization problem from yielding non-trivial solutions.
% The constraints applied ensure that all weights are both positive and normalized.
% Considering that the optimization problem functions as a maximization, the solution results in samples possessing extensive projection variances being associated with elevated weights. In simpler terms, larger weights point to outliers that exhibit significant variances within the principal subspace.}

Appendix A gives the procedure for solving the optimization problem (\ref{WeightedmeanVAR}), and the solution is
\begin{align}
\label{ai}
    a_i = \frac{\textup{exp}(\frac{1}{n\tau_a} \|\boldsymbol{P}^T(\boldsymbol{x}_i-\boldsymbol{m})\|_2^2)}{\sum_{j=1}^n\textup{exp}(\frac{1}{n\tau_a} \|\boldsymbol{P}^T(\boldsymbol{x}_j-\boldsymbol{m})\|_2^2)}.
\end{align}

The second category of outliers comprises samples within the orthogonal complement subspace. 
This is characterized by significant reconstruction errors within the principal subspace. 
Based on this observation, we proposed to solve the following optimization problem to formulate another set of weights:
% \textcolor{red}{The next category of outliers comprises samples situated within the orthogonal complement subspace. This category is identifiable by significant reconstruction errors within the principal subspace.
% Building upon this observation, we put forth the resolution of the subsequent optimization problem to formulate another set of weights:}

\begin{equation}
\begin{aligned}
\label{WeightedmeanREC}
&\max_{b_1,\cdots,b_n} \frac{1}{n}\sum_{i=1}^{n} b_i \|(\boldsymbol{I}-\boldsymbol{PP}^T)(\boldsymbol{x}_i-\boldsymbol{m})\|_2^2 + \tau_b\left(-\sum_{i=1}^nb_i\textup{ln}b_i\right)
\\
&s.t.\quad\sum_{i=1}^{n} b_i=1,\quad0\leq b_i\leq 1, 
\end{aligned}
\end{equation}

where $b_i$ is the weight for the $i$-th sample. 
Likewise, a large weight associates a sample with significant reconstruction error, i.e., an outlier in the orthogonal complement subspace. 
% \textcolor{red}{Likewise, a substantial weight is linked to a sample exhibiting notable reconstruction errors within the principal subspace, indicating that the sample belongs to the category of outliers situated within the orthogonal complement subspace.}
The solution to the optimization problem (\ref{WeightedmeanREC}) is
\begin{align}
\label{bi}
    b_i = \frac{\textup{exp}(\frac{1}{n\tau_b} \|(\boldsymbol{I} - \boldsymbol{PP}^T)(\boldsymbol{x}_i-\boldsymbol{m})\|_2^2)}{\sum_{j=1}^n\textup{exp}(\frac{1}{n\tau_b} \|(\boldsymbol{I} - \boldsymbol{PP}^T)(\boldsymbol{x}_j-\boldsymbol{m})\|_2^2)}.
\end{align}

The third category of outliers lies within both the principal component subspace and the orthogonal complement. 
%Still neither the variances nor the reconstruction errors in the principal components subspace are large enough to be distinguished. 
In this scenario, the distances from the data mean prove to be reliable indicators for these outliers.
Consequently, we proposed to solve the following optimization problem to get a set of weights for all samples:
% \textcolor{red}{The third classification of outliers exists within both the principal components subspace and its corresponding orthogonal complements.
%While the variances and reconstruction errors within the principal components subspace might not be notably large to serve as distinguishing factors,
% In this scenario, the distances from the data mean prove to be reliable indicators for these outliers.
% Consequently, we put forward the resolution of the subsequent optimization problem as a method to obtain a set of weights for all samples.}
\begin{equation}
\begin{aligned}
\label{ci}
&\max_{c_1,c_2,\cdots,c_n} \frac{1}{n}\sum_{i=1}^{n} c_i \|\boldsymbol{x}_i-\boldsymbol{m}\|_2^2+\tau_c\left(-\sum_{i=1}^{n}c_i \textup{ln}c_i\right)\\
&s.t.\quad\sum_{i=1}^{n} c_i=1,\quad0\leq c_i\leq 1,
\end{aligned}
\end{equation}
where $c_i$ is a weight for the $i$-th sample.
Similarly, a large weight implies that the corresponding sample is far from the data mean.
The solution to the optimization problem (\ref{ci}) is
\begin{align}
    c_i = \frac{\textup{exp}(\frac{1}{n\tau_c} \|\boldsymbol{x}_i-\boldsymbol{m}\|_2^2)}{\sum_{j=1}^n\textup{exp}(\frac{1}{n\tau_c} \|\boldsymbol{x}_j-\boldsymbol{m}\|_2^2)}.
\end{align}

After obtaining three sets of weights, namely $\left\{a_i\right\}_{i=1}^n, \left\{b_i\right\}_{i=1}^n, \left\{c_i\right\}_{i=1}^n$, one needs to merge them into a single set of weights, with one sample assigned one weight. 
We propose to create this final weight set as follows:
\begin{align}
\label{w_i}
    w_i = \frac{(a_ib_ic_i)^{-1}}{\sum_{j=1}^n (a_jb_jc_j)^{-1}}.
\end{align}

Since $w_i$ is determined by problem (\ref{WeightedmeanVAR}), (\ref{WeightedmeanREC}) and (\ref{ci}), formula (\ref{w_i}) considers the possibility that a sample belongs to outliers from three perspectives: principal component subspace, orthogonal complement subspace and the distance from a sample to the data center. In this way, if a sample seriously deviates from the normal samples from one of the three perspectives, it is more likely to be an outlier. Specifically, for outliers, $w_i$ tends to be small. This is because as explained earlier, at least one of the values $a_i$, $b_i$, or $c_i$ is relatively large for outliers, and consequently, $w_i$ in (\ref{w_i}) ends with a small value. Conversely, for normal samples, $w_i$ tends to be large. This is because the values of $a_i$, $b_i$, and $c_i$ are generally small for normal samples, and then, $w_i$ ends with a large value. Therefore, the formula (\ref{w_i}) distinguishes outliers from normal samples based on more essential characteristics, so that the data set has better separability, resulting in the outliers and the normal samples being identified more accurately. Formula (\ref{w_i}) effectively suppresses the influence of outliers since $w_i$ emphasizes the normal samples and makes them dominate, meanwhile minimizing the influence of outliers. At the same time, due to the continuity of the calculation of formula (\ref{w_i}), $w_i$ can more uniformly and accurately reflect the possibility that a sample belongs to outliers. Overall, the robustness of the data representation is improved.

\subsection{Estimating the sample mean and the projection matrix}

By using the weights computed in (\ref{w_i}), the PCA projection matrix as well as the sample mean can be estimated by solving the following optimization problem
\begin{align}
\label{optivariance}
&\max_{\boldsymbol{P},\boldsymbol{m}} \sum_{i=1}^{n} w_i \|\boldsymbol{P}^T(\boldsymbol{x}_i-\boldsymbol{m})\|_2^2\quad s.t.~ \boldsymbol{P}^T\boldsymbol{P} = \boldsymbol{I},
\end{align}
which maximizes the weighted sum of projection variance. 
One can also solve the following optimization problem
\begin{align}
\label{optierror}
\min_{\boldsymbol{P},\boldsymbol{m}} \sum_{i=1}^{n} w_i \|(\boldsymbol{I} - \boldsymbol{PP}^T)(\boldsymbol{x}_i-\boldsymbol{m})\|_2^2\quad s.t.~ \boldsymbol{P}^T\boldsymbol{P} = \boldsymbol{I},
\end{align}
which minimizes the weighted sum of reconstruction errors. 
In the sequel, the block coordinate descent algorithm\cite{nocedal1999numerical} is used to solve the optimization problems.
When the mean $\boldsymbol{m}$ is available, problems in (\ref{optivariance}) and (\ref{optierror}) are equivalent, since the sum of the two objective functions is
$ \sum_{i=1}^{n} w_i \|\boldsymbol{x}_i-\boldsymbol{m}\|_2^2
$, which is constant. 
Thus, $\boldsymbol{P}$ could be optimized by solving either (\ref{optivariance}) or (\ref{optierror}). 
The optimization of $\boldsymbol{m}$, on the other hand, depends on the specific optimization problem, but we will address it in the sequel.

If not otherwise specified, we focus on the optimization problem (\ref{optivariance}) to estimate $\boldsymbol{P}$.
As mentioned above, outliers are with small weights, and normal samples are with large weights.
This implies that within the objective function of (\ref{optivariance}), the projection variances of outliers are assigned small weights, while those of normal samples are with large weight coefficients.
As a result, the influence of outliers on parameter estimation is mitigated.
% \textcolor{red}{Consequently, the optimization of $\boldsymbol{P}$ can be achieved through the resolution of either (\ref{optivariance}) or (\ref{optierror}). The determination of $\boldsymbol{m}$ hinges on the specific optimization problem, which we will address subsequently.
% Unless otherwise specified, our focus remains on tackling the optimization problem (\ref{optivariance}) in order to estimate $\boldsymbol{P}$.
% As previously mentioned, outliers are associated with lower weights, while normal samples receive higher weights. This implies that within the objective function of (\ref{optivariance}), the projection variances of outliers are assigned smaller weights, while those of normal samples carry larger weight coefficients. Consequently, the impact of outliers on parameter estimation is mitigated.}

We propose to solve the problem (\ref{optivariance}) by the coordinate descent method.
In each iteration, we firstly optimize $\boldsymbol{m}$ while keeping $\boldsymbol{P}$ fixed, and then optimize $\boldsymbol{P}$ while keeping $\boldsymbol{m}$ fixed. 
% In the sequel, optimization of $\boldsymbol{m}$ and $\boldsymbol{P}$ are introduced.

% \textcolor{red}{We suggest addressing the (\ref{optivariance}) problem through the coordinate descent method. In each iteration, we begin by optimizing $\boldsymbol{m}$ while maintaining $\boldsymbol{P}$ as a fixed value. Subsequently, we optimize $\boldsymbol{P}$ while preserving $\boldsymbol{m}$ as a constant.
% Following this, the sequential optimization of both $\boldsymbol{m}$ and $\boldsymbol{P}$ will be introduced in the upcoming sections.}

When only optimizing w.r.t. $\boldsymbol{m}$ with $\boldsymbol{P}$ fixed, the optimization problem is
\begin{align}
\label{optim}
&\max_{\boldsymbol{m}} \sum_{i=1}^{n} w_i \|\boldsymbol{P}^T(\boldsymbol{x}_i-\boldsymbol{m})\|_2^2,
\end{align}
which is an unconstrained optimization problem. Taking the derivative of the objective function in (\ref{optim}) w.r.t. $\boldsymbol{m}$, and setting it to zero. There is
\begin{align}
\label{expand}
\sum_{i=1}^nw_i(-2\boldsymbol{PP}^T\boldsymbol{x_i}+2\boldsymbol{PP}^T\boldsymbol{m})=\boldsymbol{0},
\end{align}
which could be rearranged as
\begin{align}
\label{DerivativeOfMinFunc}
\boldsymbol{PP}^T(\boldsymbol{X}-\boldsymbol{m}\boldsymbol{1}^T)\boldsymbol{w}=\boldsymbol{0},
\end{align}
where $\boldsymbol{1}=[1, 1,\cdots,1]^T\in \mathbb{R}^{n\times 1}$, $\boldsymbol{w}=[w_1, w_2,\cdots,w_n]^T\in \mathbb{R}^{n\times 1}$ and $\boldsymbol{0}=[0, 0,\cdots,0]^T\in \mathbb{R}^{d\times 1}$. Since $\boldsymbol{PP}^T$ is not guaranteed to be invertible, $\boldsymbol{m}$ cannot be directly obtained by inverting $\boldsymbol{PP}^T$. 
Suppose that $(\boldsymbol{X}-\boldsymbol{m1}^T)\boldsymbol{w}$ can be decomposed into two parts: one within the subspace spanned by the columns of $\boldsymbol{P}$ and the other within its orthogonal complement subspace:
\begin{equation}
\begin{aligned}
\label{alphabetaSetting}
(\boldsymbol{X}-\boldsymbol{m1}^T)\boldsymbol{w}=\boldsymbol{P\alpha}+\boldsymbol{P}^\bot\boldsymbol{\beta},
\end{aligned}
\end{equation}
where $\boldsymbol{P}^\bot$ is the orthogonal complement of $\boldsymbol{P}$. Substituting (\ref{alphabetaSetting}) into (\ref{DerivativeOfMinFunc}) gives
\begin{align}
\boldsymbol{PP}^T(\boldsymbol{P\alpha} +\boldsymbol{P}^\bot\boldsymbol{\beta})=\boldsymbol{0},\\
\boldsymbol{P\alpha}+\boldsymbol{0}=\boldsymbol{0},
\end{align}
from which one can get $\boldsymbol{\alpha}=0$. Thus, (\ref{alphabetaSetting}) becomes
\begin{align}
\label{betaSetZero}
(\boldsymbol{X}-\boldsymbol{m1}^T)\boldsymbol{w}=\boldsymbol{P}^{\perp}\boldsymbol{\beta},
\end{align}
based on which one obtains
\begin{align}
\label{mbeta}
\boldsymbol{m}=\boldsymbol{Xw}-\boldsymbol{P}^{\perp}\boldsymbol{\beta},
\end{align}
because of $\boldsymbol{1}^T\boldsymbol{w} = \sum_{i=1}^n w_i=1$.

In the above expression, $\boldsymbol{\beta}$ is a free variable. Substituting $(\ref{mbeta})$ into the objective function  in (\ref{optim}), there is
\begin{align}
\sum_{i=1}^n w_i\|\boldsymbol{P}^T(\boldsymbol{x}_i-\boldsymbol{Xw}+\boldsymbol{P}^{\perp}\boldsymbol{\beta})\|_2^2
=\!\sum_{i=1}^n w_i\|\boldsymbol{P}^T(\boldsymbol{x}_i-\boldsymbol{Xw})\|_2^2,
\end{align}
due to $\boldsymbol{P}^T\boldsymbol{P}^{\perp}=\boldsymbol{0}$. 
Notably, from this expression, one can see that $\boldsymbol{\beta}$ does not affect the value of the objective function. 
%This is because the optimization problem (\ref{optim}) aims to maximize the projection variance or the projection on the subspace spanned by columns in $\boldsymbol{P}$, and $\boldsymbol{P}^{\perp}\boldsymbol{\beta}$ is in the orthogonal complement subspace. 

As previously discussed, both $\boldsymbol{P}$ and $\boldsymbol{m}$ can be also estimated by solving the optimization problem (\ref{optierror}). 
Following a similar approach, when $\boldsymbol{P}$ is fixed, the center can be optimized as
\begin{align}
\label{malpha}
\boldsymbol{m}=\boldsymbol{Xw}+\boldsymbol{P\alpha}.
\end{align}

It can also be demonstrated that the variable $\boldsymbol{\alpha}$ does not have any impact on the objective function in (\ref{optierror}). 
To ensure that the estimated center remains consistent across different optimization problems, the intersection in (\ref{malpha}) and (\ref{mbeta}) is used as the optimal mean:
\begin{align}
\label{m}
\boldsymbol{m}=\boldsymbol{Xw}=\sum_{i=1}^nw_i\boldsymbol{x}_i,
\end{align}
which is the weighted average of all samples, therefore, mitigating the adverse influence of outliers in learning the sample mean. (\ref{m}) is also the solution to the following optimization problem
\begin{align}
&\min_{\boldsymbol{m}} \sum_{i=1}^{n} w_i \|\boldsymbol{x}_i-\boldsymbol{m}\|_2^2,
\end{align}
whose objective function is the weighted average distance from the sample to the center, which is more consistent with the characteristic of ``center" when outliers exist.

Once $\boldsymbol{m}$ is available, $\boldsymbol{P}$ is optimized by solving the following optimization problem
\begin{align}
\label{optip}
&\max_{\boldsymbol{P}} \sum_{i=1}^{n} w_i \|\boldsymbol{P}^T(\boldsymbol{x}_i-\boldsymbol{m})\|_2^2\quad s.t.~ \boldsymbol{P}^T\boldsymbol{P} = \boldsymbol{I}.
\end{align}

Substituting (\ref{m}) into the objective function in (\ref{optip}) gives
\begin{align}
\notag
\sum_{i=1}^n w_i\|\boldsymbol{P}^T(\boldsymbol{x}_i-\boldsymbol{m})\|_2^2
&=\!\|\boldsymbol{P}^T(\boldsymbol{X}-\boldsymbol{Xw}\boldsymbol{1}^T)\textup{diag}(\sqrt{\boldsymbol{w})}\|_F^2\\
&=\textup{tr}(\boldsymbol{P}^T\boldsymbol{XHX}^T\boldsymbol{P})
\end{align}
with $\boldsymbol{H} = \textup{diag}(\boldsymbol{w}) - \boldsymbol{ww}^T$.
%\begin{align}
%\notag
    % \boldsymbol{H} &= (\boldsymbol{I}-\boldsymbol{w}\boldsymbol{1}^T)\textup{diag}(\boldsymbol{w})(\boldsymbol{I}-\boldsymbol{w}\boldsymbol{1}^T)^T\\
    % % &=\textup{diag}(\boldsymbol{w}) - \textup{diag}(\boldsymbol{w})\boldsymbol{1}\boldsymbol{w}^T - \boldsymbol{w1}^T\textup{diag}(\boldsymbol{w}) \\
    % % \notag
    % % &\quad+ \boldsymbol{w1}^T\textup{diag}(\boldsymbol{w})\boldsymbol{1}\boldsymbol{w}^T\\
    % % \notag
    % % &=\textup{diag}(\boldsymbol{w}) - \boldsymbol{w}\boldsymbol{w}^T - \boldsymbol{w}\boldsymbol{w}^T+ \boldsymbol{w}\boldsymbol{w}^T\\
    % &= \textup{diag}(\boldsymbol{w}) -\boldsymbol{ww}^T.
    %\boldsymbol{H}=\boldsymbol{W}-\frac{\boldsymbol{W11}^{T}\boldsymbol{W}}{\boldsymbol{1}^{T}\boldsymbol{W1}}
%\end{align}

The optimization problem (\ref{optip}) becomes
\begin{align}
\label{optip2}
\max_{\boldsymbol{P}} \textup{tr}(\boldsymbol{P}^T\boldsymbol{XHX}^T\boldsymbol{P})\quad s.t.~ \boldsymbol{P}^T\boldsymbol{P} = \boldsymbol{I}.
\end{align}

To solve this problem, columns in $\boldsymbol{P}$ are eigenvectors corresponding to the first $k$ largest eigenvalues of matrix $\boldsymbol{XHX}^T$ \cite{GENER}. Let the eigendecomposition of $\boldsymbol{XHX}^T$ be
\begin{align}
    \boldsymbol{XHX}^T = \boldsymbol{U\Lambda U}^T,
\end{align}
where columns in $\boldsymbol{U}$ are eigenvectors, and $\boldsymbol{\Lambda}$ is a diagonal matrix with diagonal eigenvalues in decreasing order. The solution to (\ref{optip}) is the matrix formed by the first $k$ columns of $\boldsymbol{U}$, i.e., 
\begin{align}
    \boldsymbol{P} = [\boldsymbol{U}_{.1}, \boldsymbol{U}_{.2}, \cdots, \boldsymbol{U}_{.k}],
\end{align}
where $\boldsymbol{U}_{.i}$ is the $i$-th column of matrix $\boldsymbol{U}$.

\subsection{Discussion}
The proposed RPCA-DSWL algorithm is listed in Algorithm \ref{alg:alg1}. 
The algorithm is iterative, and in each iteration, the data center (step 3), the projection matrix (step 6), and the weights (step 11) are updated sequentially.

\begin{algorithm}[H]
\caption{RPCA-DSWL}\label{alg:alg1}
\begin{algorithmic}[1]
\REQUIRE $\boldsymbol{X}=[\boldsymbol{x}_{1},\boldsymbol{x}_{2},\cdots,\boldsymbol{x}_{n}]\in\mathbb{R}^{d\times n}$, $k$ is the number of principal components.\hfill
\ENSURE  $\boldsymbol{P} \in \mathbb{R}^{d\times k}$.
\STATE Initialize $\boldsymbol{w}= \frac{1}{n}\boldsymbol{1}$.
\WHILE{not converge} 
\STATE Update the data center $\boldsymbol{m}=\boldsymbol{X}\boldsymbol{w}$.
\STATE Calculate $\boldsymbol{H}=\textup{diag}(\boldsymbol{w})-\boldsymbol{ww}^{T}$.
\STATE Calculate eigen-decomposition $\boldsymbol{XHX}^T = \boldsymbol{U\Lambda U}^T$.
\STATE $\boldsymbol{P} = [\boldsymbol{U}_{.1}, \boldsymbol{U}_{.2}, \cdots, \boldsymbol{U}_{.k}]$.
\FOR{$i=1$ to $n$}
\STATE Calculate $a_i = 
\frac{\textup{exp}\big((n\tau_a)^{-1} \|\boldsymbol{P}^T(\boldsymbol{x}_i-\boldsymbol{m})\|_2^2\big)}{\sum_{j=1}^n\textup{exp}\big((n\tau_a)^{-1} \|\boldsymbol{P}^T(\boldsymbol{x}_j-\boldsymbol{m})\|_2^2\big)}$.
\STATE Calculate $b_i = \frac{\textup{exp}\big((n\tau_b)^{-1} \|(\boldsymbol{I} - \boldsymbol{PP}^T)(\boldsymbol{x}_i-\boldsymbol{m})\|_2^2\big)}{\sum_{j=1}^n \textup{exp}\big((n\tau_b)^{-1} \|(\boldsymbol{I} - \boldsymbol{PP}^T)(\boldsymbol{x}_j-\boldsymbol{m})\|_2^2\big)}$.
\STATE Calculate $c_i = \frac{\textup{exp}\big((n\tau_c)^{-1} \|\boldsymbol{x}_i-\boldsymbol{m}\|_2^2\big)}{\sum_{j=1}^n\textup{exp}\big((n\tau_c)^{-1} \|\boldsymbol{x}_j-\boldsymbol{m}\|_2^2\big)}$.
\STATE Calculate $w_i = \frac{(a_ib_ic_i)^{-1}}{\sum_{j=1}^n (a_jb_jc_j)^{-1}}$.
\ENDFOR
\ENDWHILE
\end{algorithmic}
\label{alg1}
\end{algorithm}

RPCA-DSWL tackles the outlier issue by adaptively learning sample weights.
Similar to the proposed algorithm, PCA-DI \cite{PCADI} is also a sample weight learning method, but with a different learning strategy.
Several other robust PCA algorithms implicitly adopt the sample weight learning method. 
For instance, $L_{2p}$ PCA \cite{L2PPCA} solves the optimization by iteratively minimizing
\begin{align}
\min_{\boldsymbol{P}}\sum_{i=1}^n w_i \|\boldsymbol{x}_i-\boldsymbol{PP}^T\boldsymbol{x}_i\|_2^2,
\end{align}
where $w_i=\|\boldsymbol{x}_i-\boldsymbol{PP}^T\boldsymbol{x}_i\|_2^{p-2}$ implicitly acts as a weight.
Similarly, RPCA-OM \cite{PCAOM} estimates the data mean and the PCA projection matrix by minimizing
\begin{align}
\min_{\boldsymbol{m},\boldsymbol{P}} \sum_{i=1}^n w_i \|(\boldsymbol{I}-\boldsymbol{PP}^T)(\boldsymbol{x}_i-\boldsymbol{m})\|_2^2\quad s.t.~\boldsymbol{P}^T\boldsymbol{P} = \boldsymbol{I},
\end{align}
where $w_i=\|(\boldsymbol{I}-\boldsymbol{PP}^T)(\boldsymbol{x}_i-\boldsymbol{m})\|_2^{-1}$ serves as a weight. 

Existing sample weight learning methods typically focus on learning weights in the principal subspace or its orthogonal complement subspace, or both. 
In contrast, the RPCA-DSWL algorithm learns sample weights in three different subspaces: the original feature subspace, the principal subspace, and the orthogonal complement subspace. 
This enables the RPCA-DSWL algorithm to effectively identify outliers in various scenarios.
Furthermore, the RPCA-DSWL algorithm learns the sample mean simultaneously.

The computational complexity of the RPCA-DSWL algorithm is $O(nd^2)$. 
Step 3 updates the mean, and its computational complexity is $O(dn)$. 
The computational complexity of step 4 and step 5 is $O(nd^2)$. 
Updating sample weights from step 7 to step 11 involves computational complexity $O(kd)$. 
Therefore, the overall complexity of the proposed algorithm is $O(nd^2)$.

\section{Experiments}
In this section, we conduct extensive experiments to compare the performance of the proposed RPCA-DSWL algorithm with other robust PCA algorithms on both artificial toy datasets and real-world datasets.
The robust PCA algorithms involved in comparison are:

1) PCA based on $L_1$-norm maximization (PCA-$L_1$)\cite{PCAL1}: it designs the PCA projection matrix by solving the optimization problem (\ref{PCA-L1}), where $L_1$-norm is used as the metric for the projection variances.

2) PCA based on $R_1$-norm minimization ($R_1$-PCA)\cite{R1PCA}: $L_{2,1}$-norm of the reconstruction errors is used as the metric for designing the projection matrix.

3) PCA based on $L_{2,p}$-norm minimization ($L_{2,p}$-PCA) \cite{L2PPCA}:  $L_{2,p}$-norm of the reconstruction errors is used as the metric shown in
(\ref{PCALP}).

4) Robust PCA based on Discriminant Information (RPCA-DI)\cite{PCADI}: each sample is assigned a learned weight for discriminating outliers from normal samples. The weights can be interpreted as signal-to-noise ratio (SNR), where normal samples are with high SNR while outliers are with low SNR.
 
5) Robust PCA with Optimal Mean(RPCA-OM) \cite{PCAOM}: it learns a PCA projection matrix by using $L_{2,1}$-norm metric and simultaneously estimating the sample mean.

\subsection{Experiments on artificial toy data}
% In this part, we compare the performance of the proposed algorithm with other algorithms using manually created data.
% We created 200 two dimensional samples with zero mean and 10 outliers, the sample covariance matrix shown as follows:
We created a dataset comprising 200 two-dimensional samples to represent normal samples.
Each dimension of these samples follows a standard normal distribution.
The covariance between the two dimensions is 0.95. 
Additionally, three categories of outliers are also created.
% \begin{align}
% \left[\begin{array}{cc}
%      1&0.95  \\
%      0.95&1 
% \end{array}\right]
% \end{align}

\begin{figure*}[htbp]
\subfigure[No Outliers]
{
    \includegraphics[width=4.5cm,height=3.4cm]{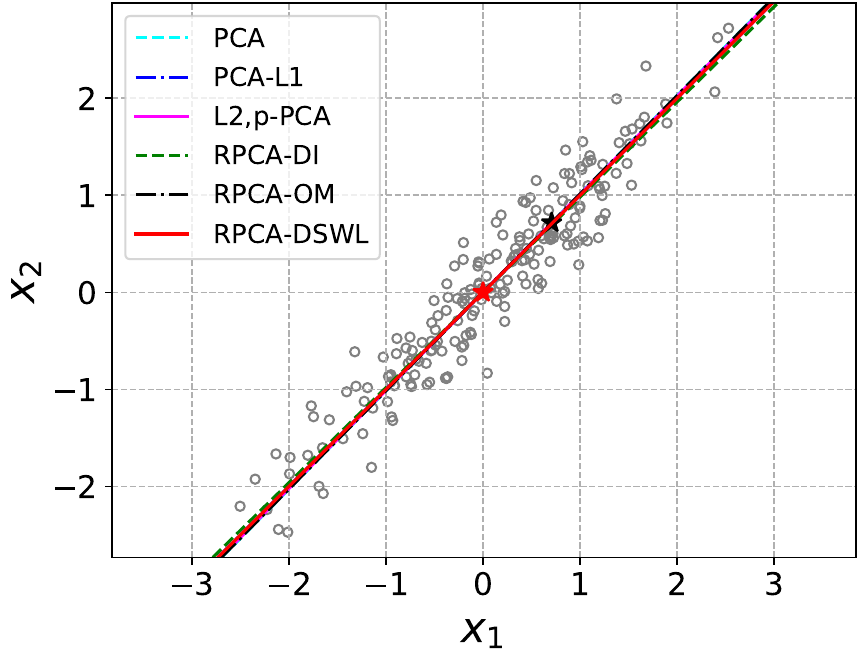}
    \hspace{-5mm}
}
\subfigure[Outliers in the PCS]
{
    \includegraphics[width=4.5cm,height=3.4cm]{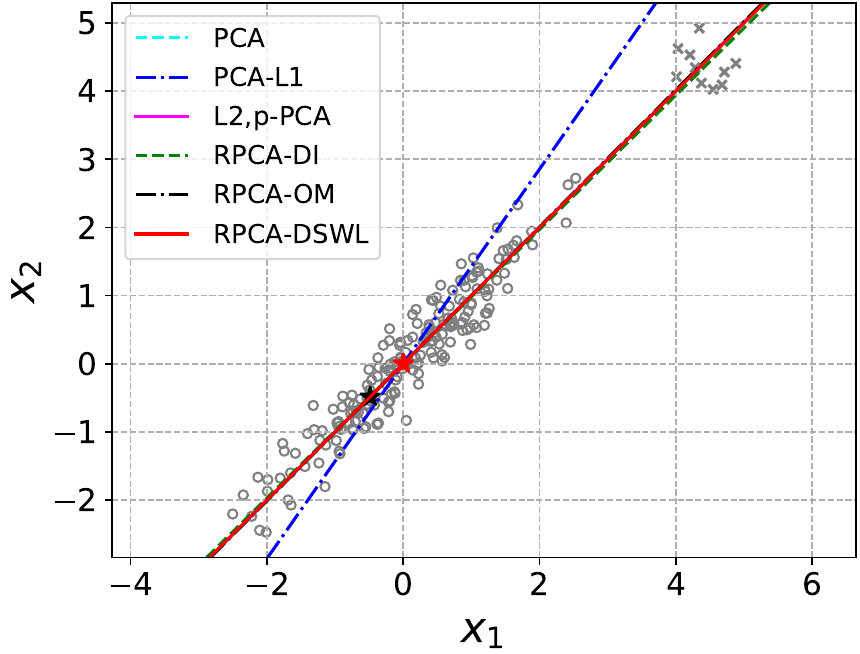}
    \hspace{-5mm}
}
\subfigure[Outliers in the OCS]
{
    \includegraphics[width=4.5cm,height=3.4cm]{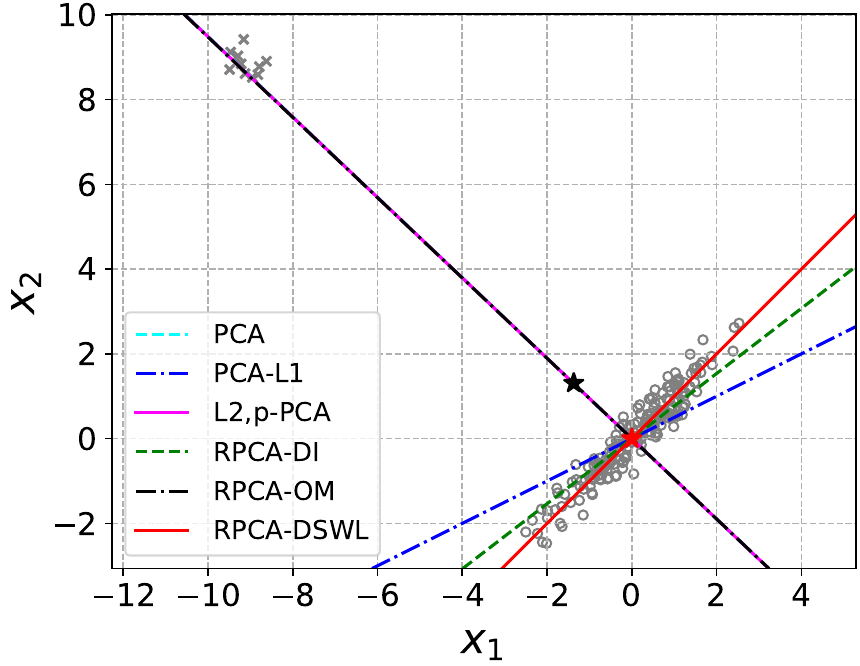}
    \hspace{-5mm}
}
\subfigure[Outliers in both PCS and OCS]
{
    \includegraphics[width=4.5cm,height=3.4cm]{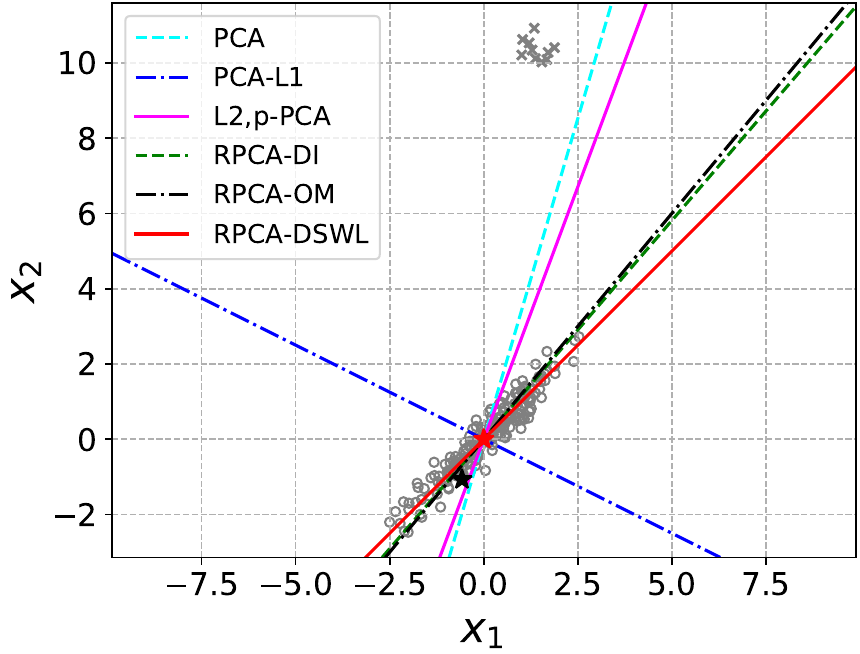}
    \hspace{-5mm}
}
\caption{The principal components extracted by six different algorithms from artificial toy datasets containing different types of outliers. $\color{gray}\circ$ represents normal sample. $\color{gray}\times$ represents outlier. $\color{black}\star$ represents the mean given by the RPCA-OM algorithm.
$\color{red}\star$ represents the mean given by the RPCA-DSWL algorithm.}
\label{figArti}
\end{figure*}

Fig. \ref{figArti} demonstrates the created samples along with principal components estimated by different algorithms ($k=1$).
In Fig. \ref{figArti}(a), where no outliers are present, it is evident that principal components produced by all algorithms are equally well and almost overlap.

Fig. \ref{figArti}(b), Fig. \ref{figArti}(c), and Fig. \ref{figArti}(d) show three categories of outliers: outliers within the principal subspace, outliers within complementary subspace, and outliers within the intersecting subspace, respectively. 
In Fig. \ref{figArti}(b) where outliers lie in the principal subspace, we observe that all algorithms perform equally well, except for PCA-L1, which gives a deviated principal direction, and RPCA-OM, which produces an inaccurate estimate of the mean.

In scenarios where the second and third categories of outliers are present, the proposed RPCA-DSWL algorithm outperforms the compared algorithms in estimating both the principal component and the sample mean. 
This can be observed in Fig. \ref{figArti}(c) and Fig. \ref{figArti}(d). 
Besides RPCA-DSWL, RPCA-DI is the second-best approach in estimating the principal component, while other algorithms are all significantly affected by outliers.

\begin{figure*}[htbp]
\subfigure[$L_{2,p}$-PCA]
{
    \includegraphics[width=4.5cm,height=3.4cm]{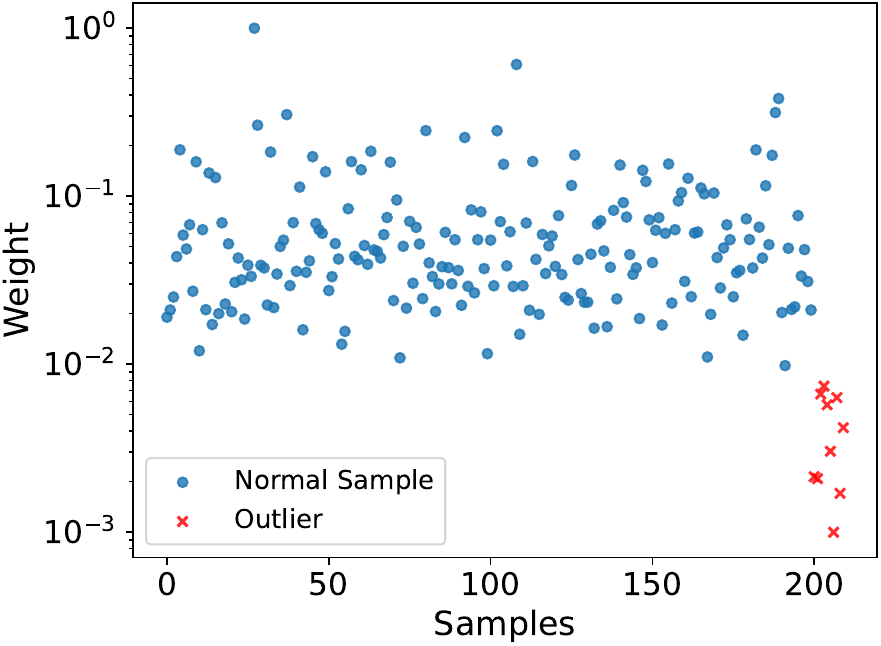}
    \hspace{-5mm}
}
\subfigure[RPCA-OM]
{

    \includegraphics[width=4.5cm,height=3.4cm]{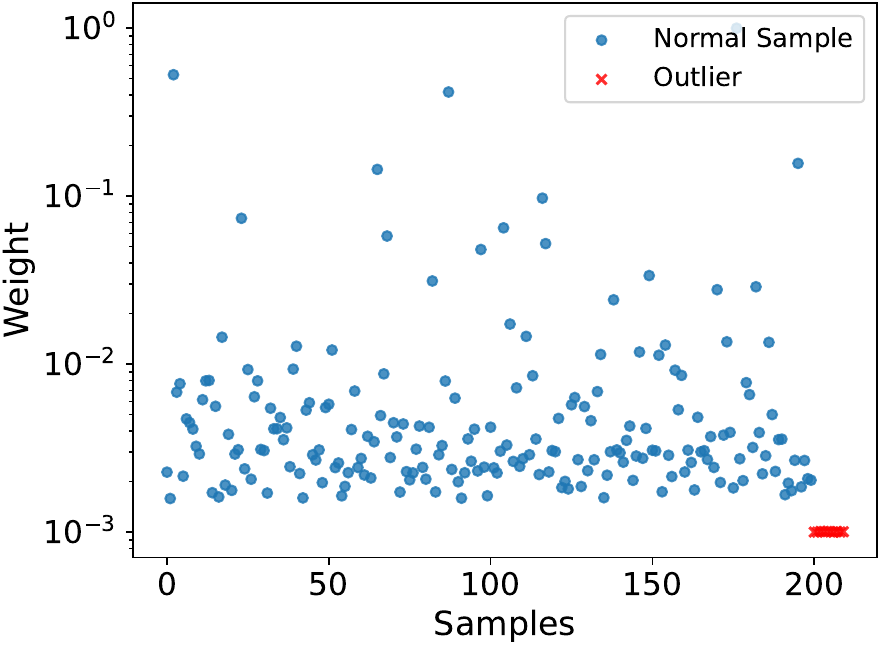}
    \hspace{-5mm}
}
\subfigure[RPCA-DI]
{
    \includegraphics[width=4.5cm,height=3.4cm]{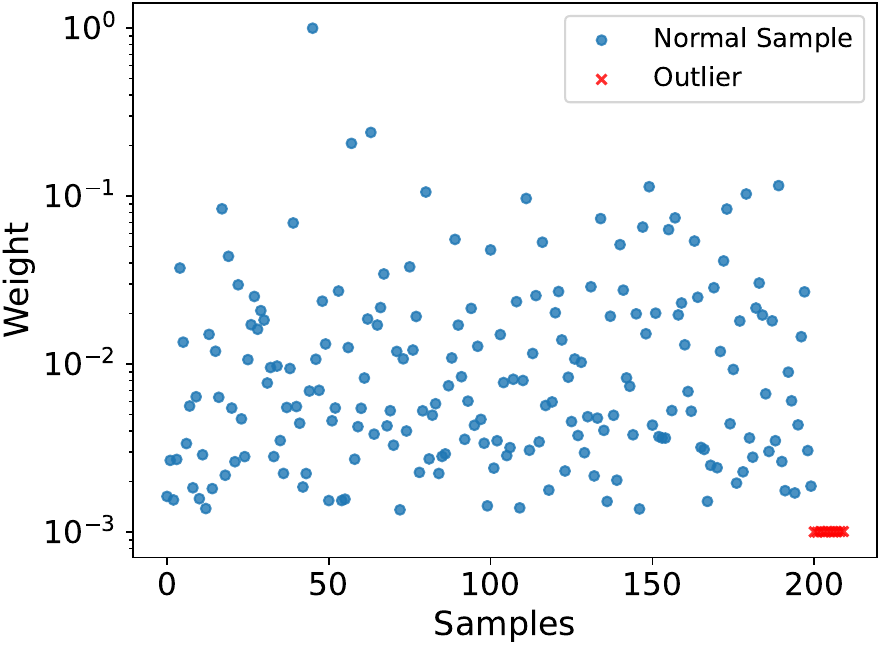}
    \hspace{-5mm}
}
\subfigure[RPCA-DSWL]
{
    \includegraphics[width=4.5cm,height=3.4cm]{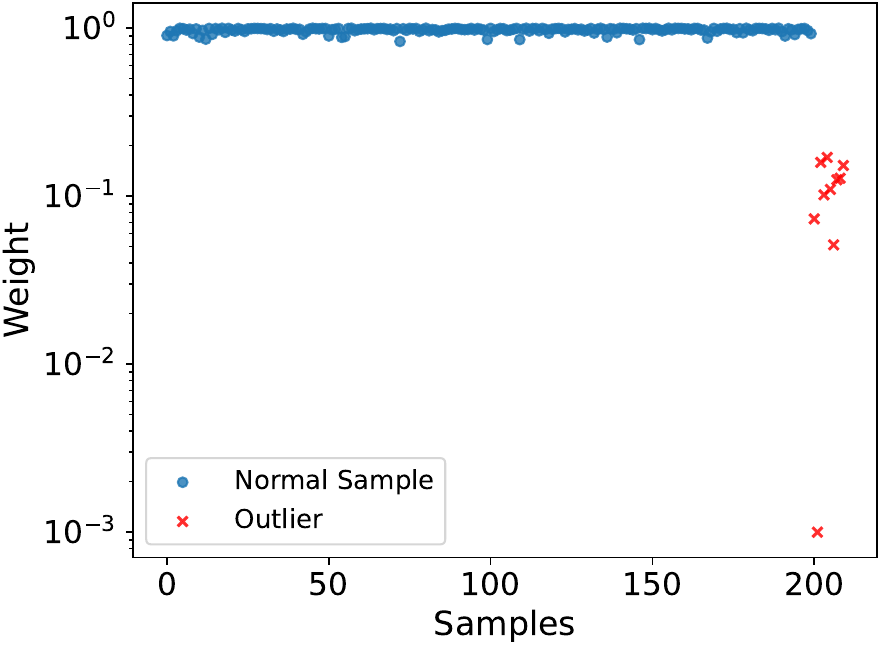}
    \hspace{-5mm}
}
\caption{From left to right, sample weights distributions produced by $L_{2,p}$-PCA, RPCA-OM, RPCA-DI, and RPCA-DSWL, respectively. For visualization purposes, the vertical axis values are in logarithmic scale.}
\label{figWeight1}
\end{figure*}

\begin{figure*}[htbp]
\subfigure[$L_{2,p}$-PCA]
{  \includegraphics[width=4.5cm,height=3.4cm]{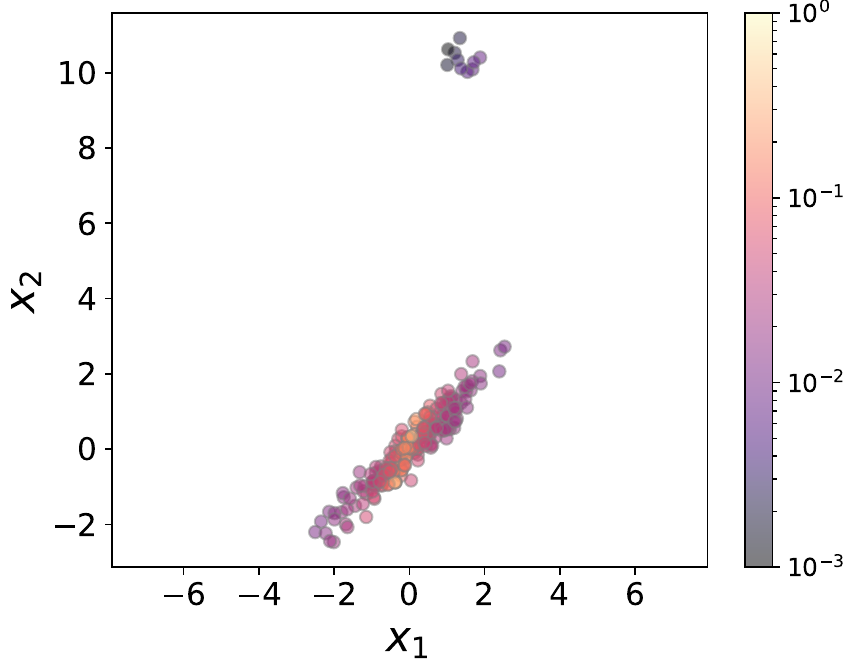}
    \hspace{-5mm}
}
\subfigure[RPCA-OM]
{   \includegraphics[width=4.5cm,height=3.4cm]{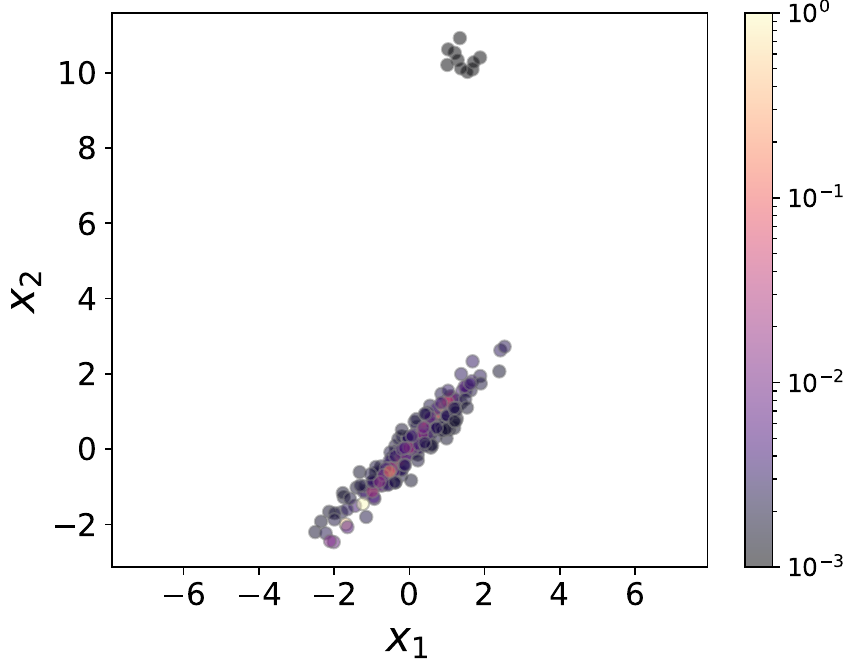}
    \hspace{-5mm}
}
\subfigure[RPCA-DI]
{ \includegraphics[width=4.5cm,height=3.4cm]{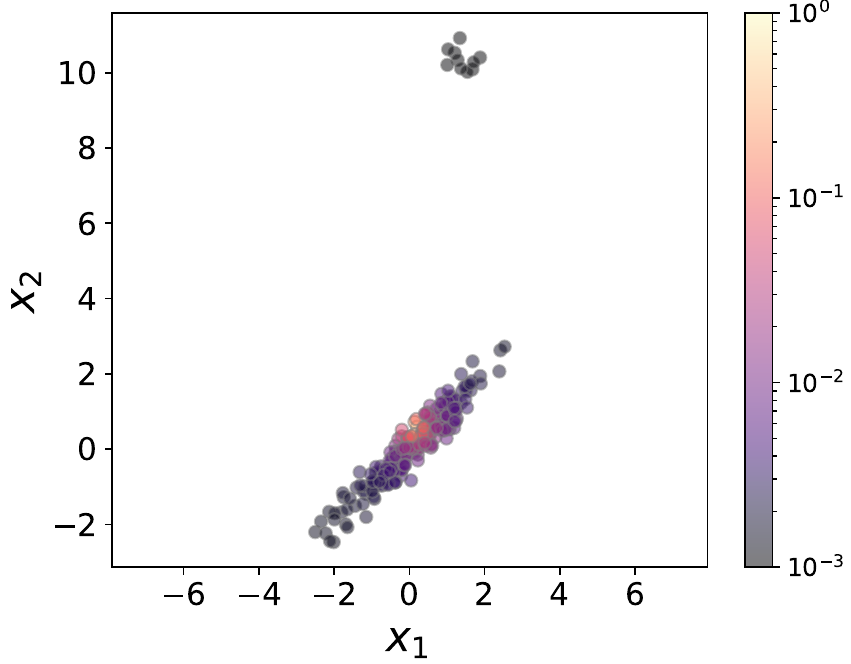}
    \hspace{-5mm}
}
\subfigure[RPCA-DSWL]
{
\includegraphics[width=4.5cm,height=3.4cm]{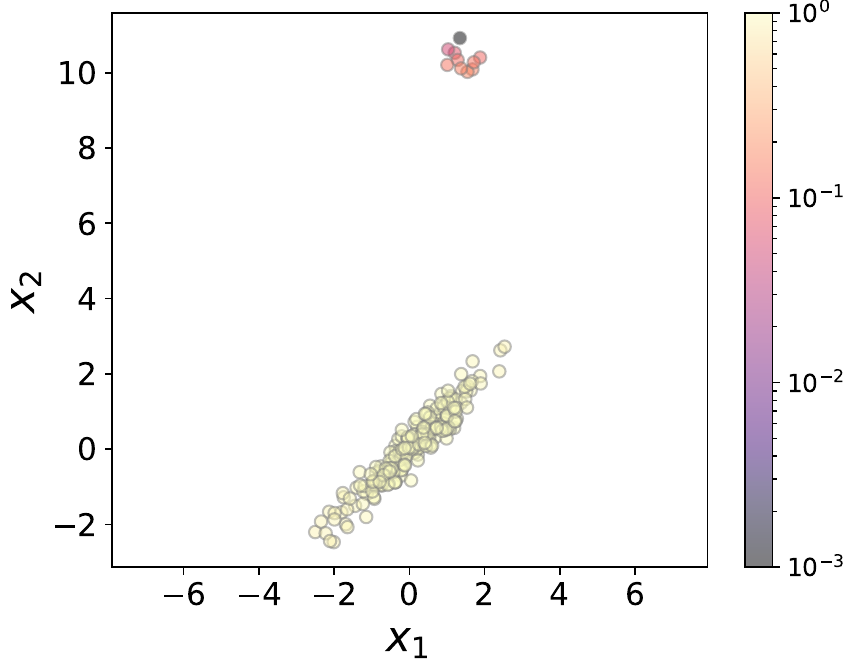}
    \hspace{-5mm}
}
\caption{Visual representation of sample weights produced by $L_{2,p}$-PCA, RPCA-OM, RPCA-DI, RPCA-DSWL (from left to right).}
\label{figWeight2}
\end{figure*}

% To further elucidate the impact of outliers on $L_{2,p}$-PCA, RPCA-DI, RPCA-OM and illustrate the robustness of RPCA-LMSW, the distribution and heat map of the weights assigned to each sample point are summarized in Fig.3 and Fig.4 when BE outliers are inserted (Since the weights calculated by RPCA-OM and RPCA-LMSW are too different in order of magnitude, the logarithm of base 10 is taken for their weights). RPCA-DI has a normalization operation divided by $ \|\boldsymbol{x}_i\|_2^2$ in the optimization process which can be treated as a part of the weight, therefore, the weight assigned to each sample by RPCA-DI can be calculated by$\frac{\lambda_i}{\|x_i\|_2^2}$ . Where $\lambda_i=\frac{u_{w,i}}{u_{\overline{w},i}}$, and $u_{w,i}$, $u_{\overline{w},i}$ can be calculated by
% \begin{equation}
% \begin{aligned}
% \label{deqn_ex23}
% \mathbf{u_w,u_{\overline{w}}}=\mathop{\arg\max}\sum_{i=1}^{n}u_{w,i}\begin{Vmatrix}\boldsymbol{P}^T\boldsymbol{x_i}\end{Vmatrix}_2
% \end{aligned}
% \end{equation}
% And the weight assigned to each sample can be calculated by 
% $\frac{1}{2\begin{Vmatrix}(\boldsymbol{I}-\boldsymbol{PP}^T)(x_i-m)\end{Vmatrix}}_2$ for RPCA-OM, and calculated by $\frac{k_i}{c_i^2}$ for RPCA-LMSW. Where $\textit{W}$ in the weight calculation process is the local optimal solution.

Fig. \ref{figWeight1} and Fig. \ref{figWeight2} depict the distributions of sample weights produced by $L_{2,p}$-PCA, RPCA-OM, RPCA-DI, and the proposed algorithm. 
In this case, we consider outliers belonging to the third category, as seen in Fig. \ref{figArti}(d). 
Both the projection variances and the reconstruction errors of these outliers are intermediate.

Fig. \ref{figWeight1} reveals that all algorithms are capable of producing discriminant weights to outliers from normal samples, namely, small weight values for outliers and large weight values for normal samples.
Nevertheless, the proposed algorithm discriminates outliers from normal samples most, as can be seen from Fig. \ref{figWeight1}(d) that the normal sample weights are greater than outlier weights by a great margin.
Although in other algorithms the normal sample weights are greater than outlier weights, the difference is not significant.

It is also evident that the normal weights in Fig. \ref{figWeight1}(d) are relatively the same, while those in Fig. \ref{figWeight1}(a)-(c) vary a lot.
We can visualize the weights for normal samples and outliers via heap maps in Fig. \ref{figWeight2}.
Fig. \ref{figWeight2}(a)-(c) demonstrate that samples located close to the data's mean are assigned large weight values, while those far from the mean receive small weights. 
This explains small weights assigned to outliers.
However, it comes at the cost of substantial variability in the weights assigned to normal samples, which distorts the pattern among the normal samples.
As a result, along with the adverse influence of outliers, the estimated principal components are inaccurate.
In contrast, the proposed algorithm gives weights with less variation, resulting in a more accurate estimation of principal components.

\subsection{Experiments on UCI datasets}

We conducted experiments on 10 datasets from the UCI dataset. 
These datasets are Seed, Ecoli, Leaf, HappinessSurvey, Zoo, Column, Breast Cancer, Glass, Wine, and BUPA.  
Table \ref{tableUCI} provides information on the number of features, categories, and samples for each dataset.
For detailed descriptions of these datasets, please refer to the UCI website\footnote{https://archive.ics.uci.edu/}.

% \begin{table}[!ht]
% \caption{UCI datasetss in the experiment  \label{tab:table1}}
% \centering
% \begin{tabular}{c c c c}
% \hline
% Name&Attributes&Categories&Sample
% \\
% \hline\hline
% Seed&7&3&207
% \\
% \hline
% Ecoli&7&8&336
% \\
% \hline
% Leaf&14&30&340
% \\
% \hline
% HappinessSurvey&6&2&143
% \\
% \hline
% Zoo&16&7&101
% \\
% \hline
% Column&6&3&310
% \\
% \hline
% Breast Cancer&30&2&569
% \\
% \hline
% Glass&9&6&214
% \\
% \hline
% Wine&13&3&178
% \\
% \hline
% BUPA&6&2&345
% \\
% \hline
% \end{tabular}
% \label{tableUCI}
% \end{table}

\begin{table*}[!ht]
\caption{Description of datasets from the UCI database used in the experiment.  \label{tab:table1}}
\centering
\begin{tabular}{c c c c c c c c c c c}
\hline
 Name of dataset& Seed & Ecoli & Leaf & HappinessSurvey & Zoo & Column & Breast Cancer & Glass & Wine & BUPA\\
\hline
Number of attributes& 7 & 7& 14 & 6 & 16 & 6 & 30 & 9 & 13 & 6\\
Number of categories& 3 & 8 & 30 & 2 & 7 & 3 & 2 & 6 & 3 & 2\\
Number of samples & 207 & 336 & 340 & 143 & 101 & 310 & 569 & 214 & 178 & 345\\
\hline
\end{tabular}
\label{tableUCI}
\end{table*}

\begin{table*}[!ht]
\setlength{\tabcolsep}{5pt}
\renewcommand{\arraystretch}{1.25}
\caption{The average accuracy of KNN using features extracted by different PCA algorithms ($\%$). \label{tab:table3}}
    \centering
    \begin{tabular}{c c c c c c c c c c c c c c c c }
    \toprule
    \multirow{2}{*}{Dataset}& \multicolumn{3}{c}{PCA}&\multicolumn{3}{c}{PCA-$L_1$}&\multicolumn{3}{c}{$R_1$-PCA}&\multicolumn{3}{c}{RPCA-DI}&\multicolumn{3}{c}{RPCA-DSWL} \\
  \cmidrule(lr){2-4}\cmidrule(lr){5-7}\cmidrule(lr){8-10}\cmidrule(lr){11-13}\cmidrule(lr){14-16}
&1&3&5 &1&3&5 &1&3&5 &1&3&5 &1&3&5\\
  \hline
      Seed & \underline{76.67} & 86.61 & 89.56 & 42.26 & \underline{87.18} & 89.46 & 71.84 & 86.81 & \underline{89.61} & 71.31 & \textbf{89.56} & \underline{89.61} & \textbf{77.11} & \textbf{89.56} & \textbf{89.7} \\ 
      Ecoli & 44.29 & 72.48 & 80.15 & 39.83 & \textbf{75} & 80.62 & 54.42 & 70.83 & 79.46 & \textbf{60.39} & 73.5 & \textbf{81.83} & \underline{56.53} & \underline{73.93} & \underline{81.07} \\ 
      Leaf & 9.26 & 48.56 & 60.41 & \textbf{11.59} & 42.74 & 59.06 & 9.85 & 49.62 & 60.59 & 7.91 & 50.18 & \textbf{65.71} & \underline{10.35} & \textbf{50.24} & \underline{64.74} \\ 
      Happiness & \underline{59.32} & \textbf{58.68} & \textbf{55.62} & 55.9 & \underline{57.88} & 55.14 & 56.86 & 58.23 & 54.29 & 56.28 & 54.02 & 54.17 & \textbf{59.84} & 54.01 & \underline{55.38} \\ 
       Zoo & 82.88 & 90.68 & 92.27 & 71.8 & 88.23 & 92.97 & 82.75 & \underline{95.82} & 94.62 & \underline{83.26} & \textbf{95.92} & 94.72 & \textbf{86.14} & \textbf{95.92} & \underline{94.92} \\ 
       Column & 63.23 & \textbf{76.9} & \underline{82.45} & 61.68 & 74.13 & 81.06 & \underline{66.13} & 73.71 & \underline{82.45} & 64.97 & 73.48 & \underline{82.45} & \textbf{66.58} & \underline{75.52} & \textbf{82.52} \\ 
        BreastCancer & 87.96 & \textbf{93.04} & 95.08 & 74.04 & 90.62 & 92.97 & 88.22 & 91.11 & 95.17 & \underline{88.52} & 90.58 & \underline{95.26} & \textbf{89.22} & \underline{92.06} & \textbf{95.57} \\ 
         Glass & 45.09 & 56.95 & 58.93 & 37.38 & 55.75 & 60 & \underline{49.47} & \underline{58.34} & 59.71 & 42.53 & 56.03 & \underline{60.51} & \textbf{50.18} &\textbf{ 59.71} & \textbf{61.06} \\ 
        Wine & 71.37 & 84.37 & \underline{86.99} & 56.51 & 81.77 & 89.11 & 69.29 & 84.78 & 86.88 & \underline{71.57} & \underline{87.67} & 89.53 & \textbf{74.24} & \textbf{88.35} & \textbf{90.89}\\ 
       BUPA & 43.25 & 97.04 & 94.93 & \textbf{57.98} & 77.21 & 92.41 & 51.55 & 75.72 & \textbf{98.78} & \underline{54.98} & \textbf{98.84} & \textbf{98.78} & 52.92 & \underline{97.23} & \underline{97.05}\\ 
%       \textbf{Average} & 58.33 & 76.53 & 79.64 & 50.90 & 73.05 & 79.28 & 60.04 & 74.50 & 80.16 & \underline{60.17} & \underline{76.98} & \underline{81.26} & \textbf{62.31} & \textbf{77.65} & \textbf{81.29} \\
\bottomrule
\end{tabular}
\label{tableUCIKNN}
\end{table*}

To assess the robustness of each algorithm, artificial contamination was added to some samples in datasets to convert them into outliers. 
We randomly selected 25\% samples from each dataset and then amplified half of the features 5 times, 10 times, or 20 times randomly. We utilized different PCA algorithms to extract features of these contaminated datasets and then used these features to train the K-Nearest Neighbor (KNN) algorithm with K being 1. 
A ten-fold cross-validation method is employed to compute the average accuracy of KNN in each dataset. 
% We normalise each training datasets and normalise the test datasets using the mean and variance of the training datasetss. 

Table \ref{tableUCIKNN} shows the average accuracy of the KNN algorithm for different numbers of extracted principal components: 1, 3, and 5. 
Within the table, the best results are highlighted in bold, while the second-best results are indicated by underlining. 
The results in table \ref{tableUCIKNN} demonstrate that the proposed RPCA-DSWL algorithm outperforms all other algorithms in most cases. 
In certain cases, the results produced by the proposed algorithm are not the best, but they are the second-best, except in two cases.
These facts show the robustness of the proposed RPCA-DSWL algorithm in estimating the principal components from datasets containing outliers.

\subsection{Experiments on face images}

We conduct experiments on the following benchmark face image datasets.

1) The YALE face dataset\footnote{\url{http://cvc.cs.yale.edu/cvc/projects/yalefaces/yalefaces.html}}, which contains  15 distinct subjects. There are 11 images per subject, one per different facial expression or configuration.

2) The extended Yale B dataset\footnote{\url{https://www.cs.yale.edu/cvc/projects/yalefacesB/yalefacesB.html} }\cite{YALE}, which contains 2414 frontal-face images over 38 subjects and about 64 images per subject. The images were captured under different lighting conditions and various facial expressions.

% it contains 400 face images of 40 distinct subjects. There are 10 images per subject. The images were taken at the Olivetti Research Laboratory in Cambridge, UK, between April 1992 and April 1994. It is a popular dataset for face recognition research. 

3) the ORL dataset\footnote{\url{https://cam-orl.co.uk/facedatabase.html}}\cite{orl}, which includes 400 images corresponding to 40 distinct individuals with 10 images for each individual. 
Images were taken at different times, in different lighting conditions, and under various facial expressions.

4) The Umist dataset\footnote{https://www.visioneng.org.uk/datasets}\cite{umist}, which consists of 564 images of 20 individuals (mixed race/gender/appearance). Each individual was photographed in a range of poses from profile to frontal views.

% 4) The extended Yale B datasbase\footnote{\url{https://www.cs.yale.edu/cvc/projects/yalefacesB/yalefacesB.html} }\cite{YALE}, which contains 2414 frontal-face images over 38 subjects and about 64 images per subject. The images were captured under different lighting conditions and various facial expressions.

% \begin{figure}[htbp]
% \centering
% \subfigure[]{
% \begin{minipage}{0.49\linewidth}
%     \centering
%     \includegraphics[width=0.1\linewidth]{Fig.4(a) Yale B.pdf}
% \end{minipage}}
% \quad

% \subfigure[]{
% \begin{minipage}{0.49\linewidth}
% 	\includegraphics[width=0.1\linewidth]{Fig.4(b) Umist.pdf}
% \end{minipage}}\quad
	
% \subfigure[]{
% \begin{minipage}{0.49\linewidth}
% 	\includegraphics[width=0.1\linewidth]{Fig.4(c) ORL.pdf}
% \end{minipage}}\quad
% \subfigure[]{
% \begin{minipage}{0.49\linewidth}
% 	\includegraphics[width=0.1\linewidth]{Fig.4(d) Olivetti.pdf}
% \end{minipage}}
% \end{figure}

\begin{figure*}
\centering
\includegraphics[width=5in]{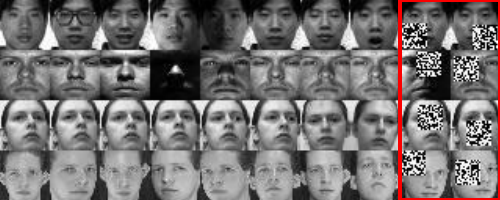}
\caption{
Sampled images from different datasets. From top to bottom, the datasets sampled are Yale, Extended Yale B, ORL, and Umist, respectively. 
Images in the red box are with random black and white blocks, representing manually created outlier images.}
\label{figFaceImage}
\end{figure*}

\begin{figure*}
\centering
\subfigure[Yale]
{
\includegraphics[width=4.5cm,height=3.4cm]{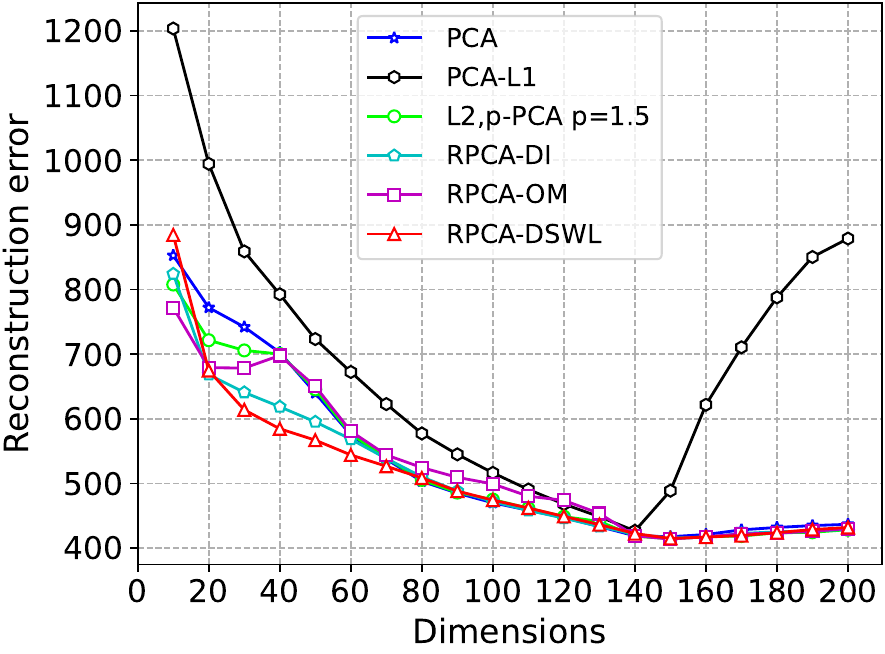}
\hspace{-3.8mm}}
\subfigure[Extended Yale B]{
\includegraphics[width=4.5cm,height=3.4cm]{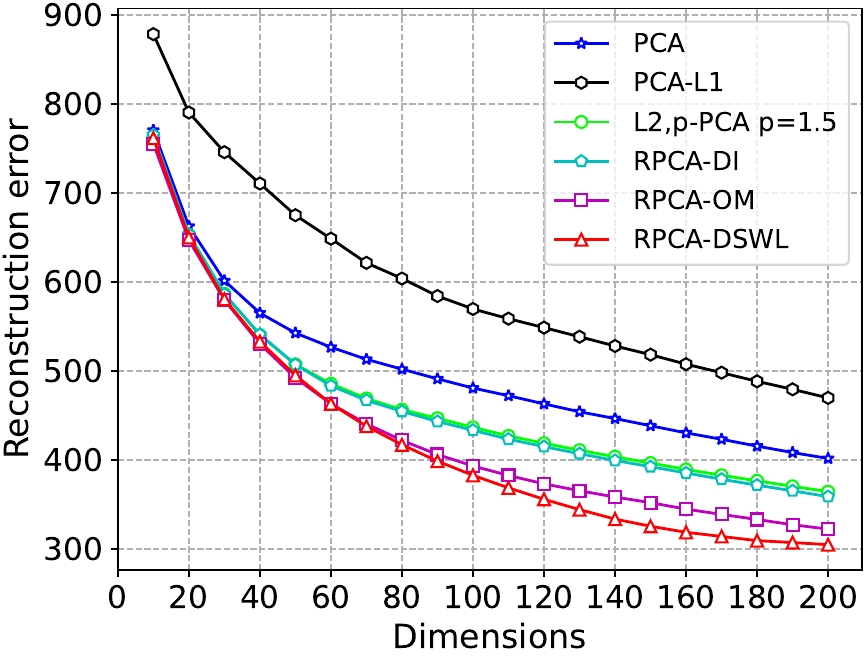}
\hspace{-3.8mm}}
\subfigure[ORL]{
\includegraphics[width=4.5cm,height=3.4cm]{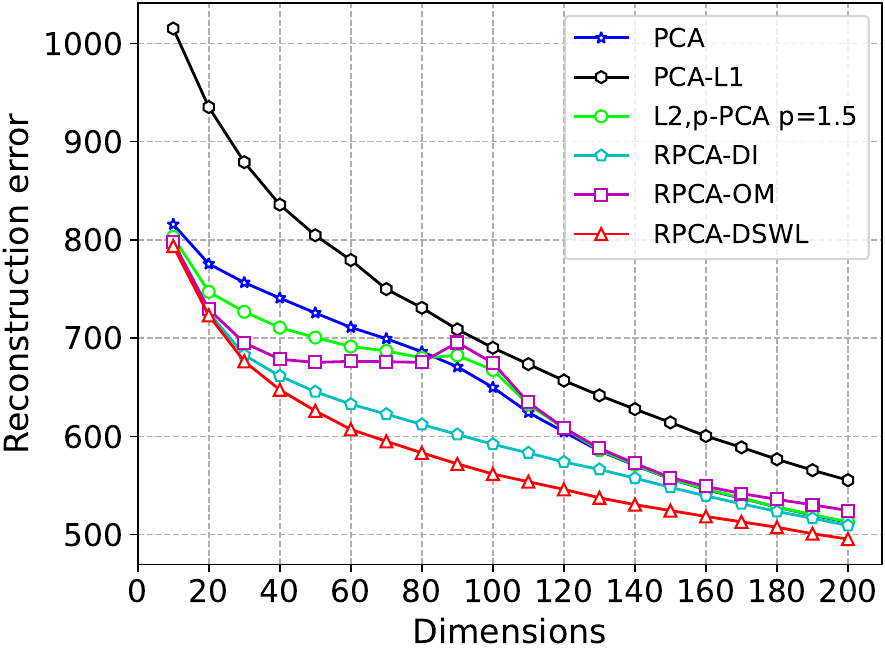}
\hspace{-3.8mm}}
\subfigure[Umist]{
\includegraphics[width=4.5cm,height=3.4cm]{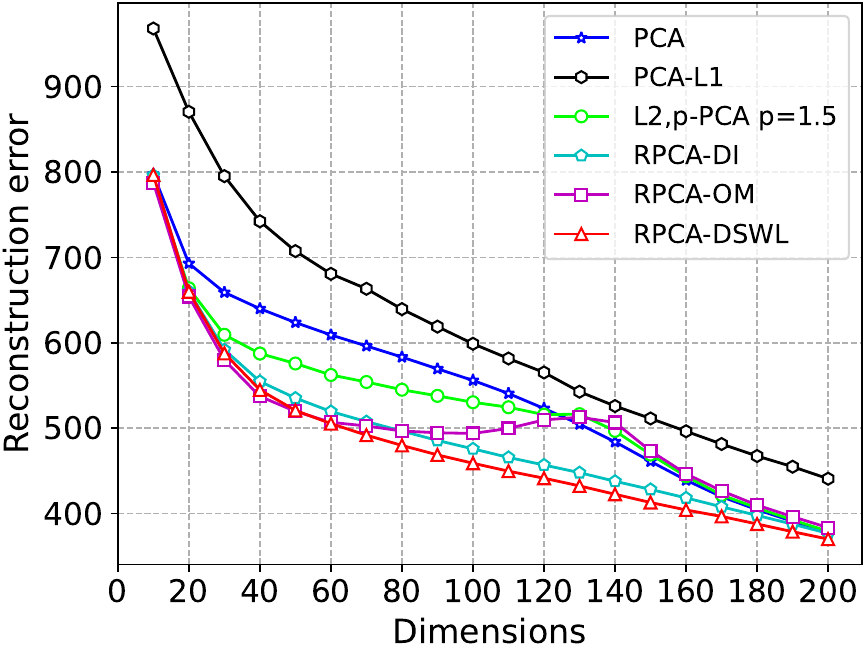}
\hspace{-3.8mm}}
\centering
\caption{The average reconstruction errors of images from different datasets produced by the six algorithms. Datasets, from left to right, are Yale, extended Yale B, ORL, and Umist datasets, respectively.}
\label{figError}
\end{figure*}

\begin{figure*}
\centering
\subfigure[$L_{2,p}$-PCA]{
\includegraphics[width=4.5cm,height=3.4cm]{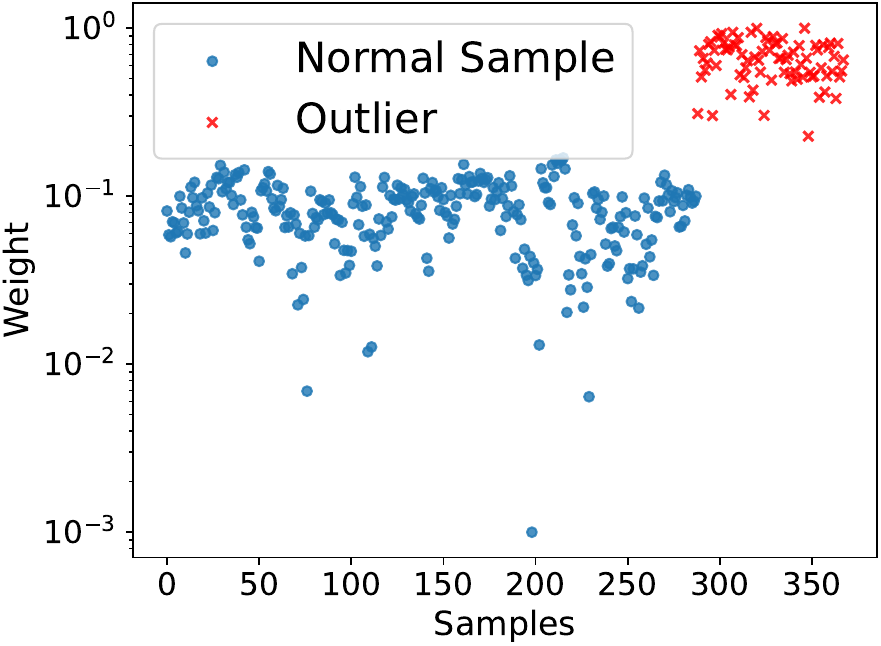}}
\hspace{-3.8mm}
\subfigure[RPCA-OM]{
\includegraphics[width=4.5cm,height=3.4cm]{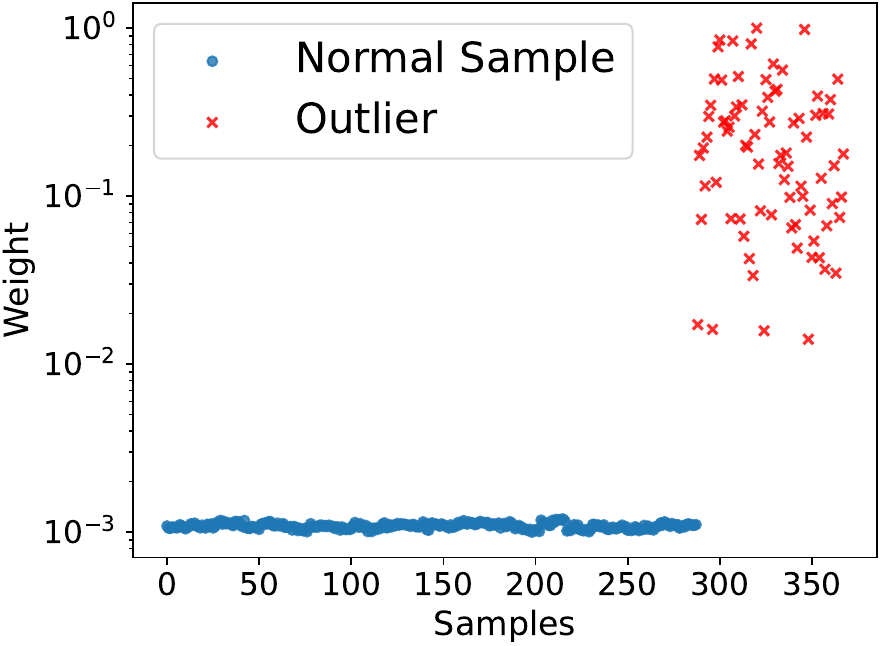}}
\hspace{-3.8mm}
\subfigure[RPCA-DI]{
\includegraphics[width=4.5cm,height=3.4cm]{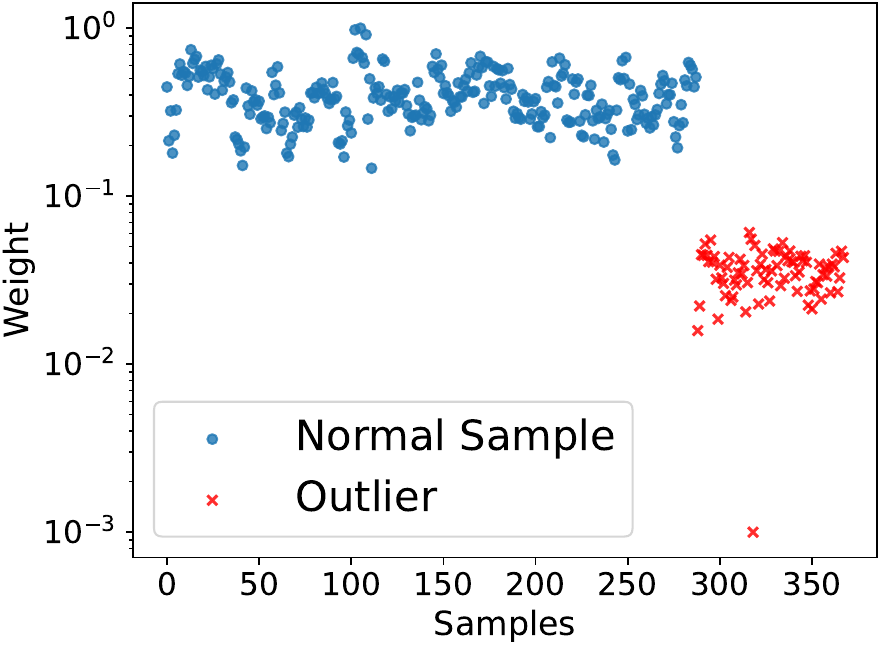}}
\hspace{-3.8mm}
\subfigure[RPCA-DSWL]{
\includegraphics[width=4.5cm,height=3.4cm]{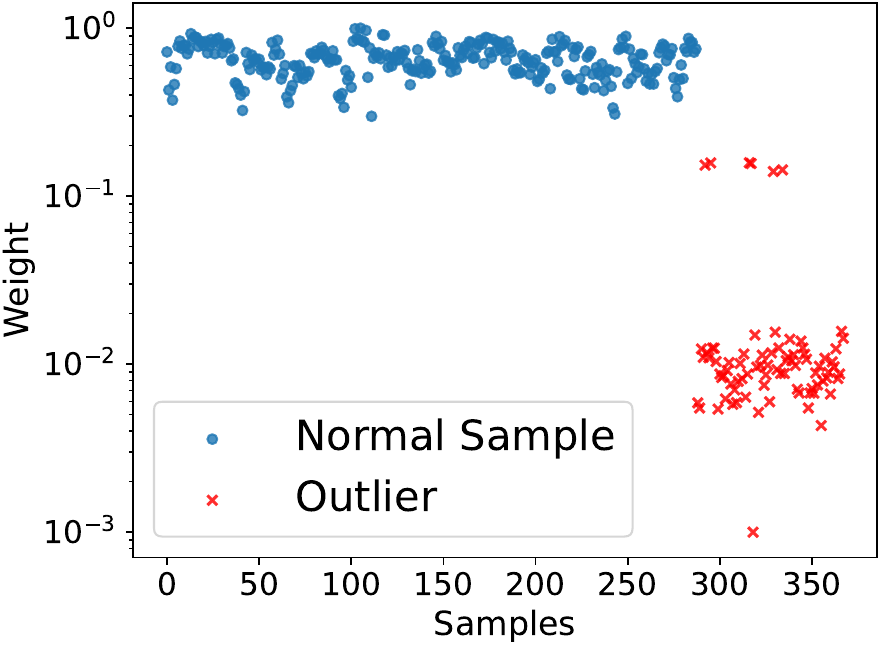}
\hspace{-3.8mm}}
\label{fig:subfig}
\centering
\caption{Using the ORL dataset, distributions of sample weights produced by $L_{2,p}$-PCA, RPCA-OM, RPCA-DI and RPCA-DSWL, respectively (from left to right). For visualization purposes, the vertical axis values are in logarithmic scale.}
\label{figFaceWeight}
\end{figure*}

\begin{figure*}
  \begin{tabular}{  c   c  c  c  c  c  c  c }
    \hline
    datasetss & Original & PCA & PCA-L1 & RPCA-OM & $L_{2,p}$-PCA & RPCA-DI & RPCA-DSWL \\ \hline\hline
    YALE &
    \begin{minipage}[b]{0.2\columnwidth}
		\centering
		\raisebox{-0.5\height}
  {\includegraphics[width=0.75\linewidth]{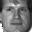}}
    
	\end{minipage}
    & 
    \begin{minipage}[b]{0.2\columnwidth}
		\centering
		\raisebox{-.5\height}
  {\includegraphics[width=0.75\linewidth]{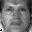}}
	\end{minipage}
    & 
    \begin{minipage}[b]{0.2\columnwidth}
		\centering
		\raisebox{-.5\height}
  {\includegraphics[width=0.75\linewidth]{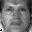}}
	\end{minipage}
	&
    \begin{minipage}[b]{0.2\columnwidth}
		\centering
		\raisebox{-.5\height}
  {\includegraphics[width=0.75\linewidth]{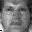}}
	\end{minipage}
	&
		 \begin{minipage}[b]{0.2\columnwidth}
		\centering
		\raisebox{-.5\height}
  {\includegraphics[width=0.75\linewidth]{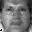}}
	\end{minipage}
 	&
		 \begin{minipage}[b]{0.2\columnwidth}
		\centering
		\raisebox{-.5\height}
  {\includegraphics[width=0.75\linewidth]{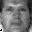}}
	\end{minipage}
	&
		 \begin{minipage}[b]{0.2\columnwidth}
		\centering
		\raisebox{-0.5\height}
  {\includegraphics[width=0.75\linewidth]{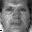}}
	\end{minipage}\\
& &\small 21.25 dB&21.24 dB&\underline{21.47} dB&21.21dB&21.24 dB&\textbf{21.92} dB
    \\ 

        Exteded Yale B &
    \begin{minipage}[b]{0.2\columnwidth}
		\centering
		\raisebox{-.5\height}
  {\includegraphics[width=0.75\linewidth]{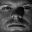}}
	\end{minipage}
    & 
    \begin{minipage}[b]{0.2\columnwidth}
		\centering
		\raisebox{-.5\height}
  {\includegraphics[width=0.75\linewidth]{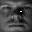}}
	\end{minipage}
    & 
    \begin{minipage}[b]{0.2\columnwidth}
		\centering
		\raisebox{-.5\height}
  {\includegraphics[width=0.75\linewidth]{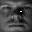}}
	\end{minipage}
	&
    \begin{minipage}[b]{0.2\columnwidth}
		\centering
		\raisebox{-.5\height}
  {\includegraphics[width=0.75\linewidth]{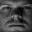}}
	\end{minipage}
	&
		 \begin{minipage}[b]{0.2\columnwidth}
		\centering
		\raisebox{-.5\height}
  {\includegraphics[width=0.75\linewidth]{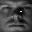}}
	\end{minipage}
 	&
		 \begin{minipage}[b]{0.2\columnwidth}
		\centering
		\raisebox{-.5\height}
  {\includegraphics[width=0.75\linewidth]{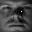}}
	\end{minipage}
	&
		 \begin{minipage}[b]{0.2\columnwidth}
		\centering
		\raisebox{-.5\height}
  {\includegraphics[width=0.75\linewidth]{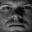}}
	\end{minipage}
    \\ 
&&24.33 dB&24.32 dB&\underline{29.55} dB&24.77 dB&26.49 dB&\textbf{30.29} dB
    \\
        ORL &
    \begin{minipage}[b]{0.2\columnwidth}
		\centering
		\raisebox{-.5\height}
  {\includegraphics[width=0.75\linewidth]{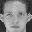}}
	\end{minipage}
    & 
    \begin{minipage}[b]{0.2\columnwidth}
		\centering
		\raisebox{-.5\height}
  {\includegraphics[width=0.75\linewidth]{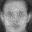}}
	\end{minipage}
    & 
    \begin{minipage}[b]{0.2\columnwidth}
		\centering
		\raisebox{-.5\height}
  {\includegraphics[width=0.75\linewidth]{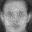}}
	\end{minipage}
	&
    \begin{minipage}[b]{0.2\columnwidth}
		\centering
		\raisebox{-.5\height}
  {\includegraphics[width=0.75\linewidth]{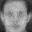}}
	\end{minipage}
	&
		 \begin{minipage}[b]{0.2\columnwidth}
		\centering
		\raisebox{-.5\height}
  {\includegraphics[width=0.75\linewidth]{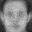}}
	\end{minipage}
 	&
		 \begin{minipage}[b]{0.2\columnwidth}
		\centering
		\raisebox{-.5\height}
  {\includegraphics[width=0.75\linewidth]{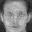}}
	\end{minipage}
	&
		 \begin{minipage}[b]{0.2\columnwidth}
		\centering
		\raisebox{-.5\height}
  {\includegraphics[width=0.75\linewidth]{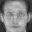}}
	\end{minipage}
    \\
    & &21.78 dB&21.78 dB&21.69 dB&21.67 dB&\underline{22.48} dB&\textbf{23.03} dB
    \\
    Umist &
    \begin{minipage}[b]{0.2\columnwidth}
		\centering
		\raisebox{-.5\height}
  {\includegraphics[width=0.75\linewidth]{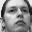}}
	\end{minipage}
    & 
    \begin{minipage}[b]{0.2\columnwidth}
		\centering
		\raisebox{-.5\height}
  {\includegraphics[width=0.75\linewidth]{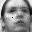}}
	\end{minipage}
    & 
    \begin{minipage}[b]{0.2\columnwidth}
		\centering
		\raisebox{-.5\height}
  {\includegraphics[width=0.75\linewidth]{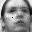}}
	\end{minipage}
	&
    \begin{minipage}[b]{0.2\columnwidth}
		\centering
		\raisebox{-.5\height}
  {\includegraphics[width=0.75\linewidth]{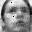}}
	\end{minipage}
	&
		 \begin{minipage}[b]{0.2\columnwidth}
		\centering
		\raisebox{-.5\height}
  {\includegraphics[width=0.75\linewidth]{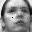}}
	\end{minipage}
 	&
		 \begin{minipage}[b]{0.2\columnwidth}
		\centering
		\raisebox{-.5\height}
  {\includegraphics[width=0.75\linewidth]{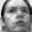}}
	\end{minipage}
	&
		 \begin{minipage}[b]{0.2\columnwidth}
		\centering
		\raisebox{-.5\height}
  {\includegraphics[width=0.75\linewidth]{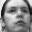}}
	\end{minipage}
    \\
 & &22.92 dB&22.91 dB&21.49 dB&23.32 dB&\underline{23.43} dB&\textbf{25.96} dB
    \\
    \hline\hline
  \end{tabular}
\caption{ Reconstruction errors of images from different datasets on the principal subspace produced by PCA, PCA-$L_1$, RPCA-OM, $L_{2,p}$-PCA, RPCA-DI, and RPCA-DSWL, respectively. 
PSNRs are also shown under the reconstructed images.}
\label{figRecoveredImage}
\end{figure*}

% \begin{table*}[!ht]
% \caption{PSNR between reconstructed image and original image}
% \centering
% \begin{tabular}{c c c c c c c}
% \hline
% PSNR(dB)&PCA & PCA-L1 & RPCA-OM & $L_{2,p}$-PCA & RPCA-DI & Proposed 
% \\
% \hline
% Olivetti&20.36&20.36&19.96&20.18&\underline{20.45}&\textbf{21.82}
% \\
% ORL&21.78&21.78&21.69&21.67&\underline{22.48}&\textbf{23.03}
% \\
% Yale&24.33&24.32&\underline{29.55}&24.77&26.49&\textbf{30.29}
% \\
% Umist&22.92&22.91&21.49&23.32&\underline{23.43}&\textbf{25.96}
% \\
% \hline
% \end{tabular}
% \label{tab:table1}
% \end{table*}

% In the experiment, the size of each face image was adjusted to $\textit{32}\times\textit{32}$ pixels.
In these experiments, we resized each image to 32 by 32 to save computation time and
ensure uniformity.
All samples were normalized with zero mean, except for the RPCA-OM and the proposed RPCA-DSWL which are capable of learning the mean.

During the training phase, we randomly selected 20\% samples from the training data and manually transformed them into outliers by randomly selecting a square block within the image and filling it with black and white noises.
The size of the block is one-quarter of the image.
Fig. \ref{figFaceImage} displays face images from the four datasets, with the last two columns representing manually created outlier images.

We compared errors of reconstructing images from their projection onto principal component subspace produced by different algorithms.
The procedures of the experiment are as follows: 1) we estimated the principal components from the training data; 2) we calculated the average reconstruction errors of the testing data as follows:
% In this part, we conduct reconstruction experiments. First calculate the matrix of principal components in the training set, vary the projection dimension from 10 to 200 with an interval of 10, then calculate the average reconstruction error corresponding to each projection dimension in the testing set by
\begin{equation}
\begin{aligned}
\label{deqn_ex21}
\textit{error}=\frac{1}{n}\sum_{i=1}^{n}\|\boldsymbol{x}_i^{\textit{test}}-\boldsymbol{PP}^T\boldsymbol{x}_i^{\textit{text}}\|_2
\end{aligned}
\end{equation}
where $\boldsymbol{x}_i^{\textit{text}}$ is the $i$-th testing sample; 
3) We used a ten-fold cross-validation method to compute the average testing errors.

Fig. \ref{figError} illustrates the average reconstruction errors by different algorithms.
The dimension of the principal components varies from 10 to 200 with an interval of 10.
We can see the average reconstruction errors are reduced as the dimension grows.
Notably, the RPCA-DSWL algorithm consistently yields the smallest average reconstruction errors across the four datasets, highlighting the robustness of the proposed algorithm against outliers.
Apart from the RPCA-DSWL algorithm, RPCA-DI is the second-best-performing algorithm, except for its performance on the extended Yale B dataset.

Fig. \ref{figFaceWeight} displays the weights assigned to normal samples and outliers by $L_{2,p}$-PCA, RPCA-OM, RPCA-DI, and the proposed RPCA-DSWL algorithm. 
The dimension of the principal components is set to 100.
It can be seen that the weights produced by RPCA-DI and the proposed algorithm effectively discriminate outliers from normal samples, as outlier weights are smaller than normal sample weights.
In contrast, $L_{2, p}$-PCA and RPCA-OM do not give discriminative weights. 
Even worse, outlier weights are larger than the normal sample weights, which explains their bad performance in terms of reconstruction errors.

Comparing Fig. \ref{figFaceWeight}(c) and Fig. \ref{figFaceWeight} (d), one can see that the gaps between the normal sample weights and outlier weights by the proposed RPCA-DSWL algorithm are greater than those by RPCA-DI.
This is similar to what we previously observed that the weights produced by the proposed algorithm are better than those produced by RPCA-DI.
In other words, the proposed RPCA-DSWL algorithm has superior capability in discriminating outliers from normal samples. 

% As large gaps show significantly difference in weights for the difference between the normal sample weights and outlier weights produced by the proposed 

% Among them, the reduction of RPCA-LMSW is particularly obvious. Moreover, in the process of increasing projection dimension, the other five methods all have the phenomenon of increasing reconstruction error, indicating that the effect and order of their principal component extraction are different from the actual optimal solution. While RPCA-LMSW decreases steadily and smoothly throughout the process, which means that our algorithm is remarkably superior to the others for reconstructing data when there are outliers in the datasets and indicates that RPCA-LMSW is effective and robust to outliers.

% In order to more intuitively show the effect of the algorithms reconstructing the face picture, we have visualized them on the four face datasetss respectively in the case where the projection dimension is 100. First, the principal component is extracted from the training set, and then one face picture is selected randomly from the testing set to be reconstructed. The result of the reconstruction is finally shown in Fig.8.

Fig. \ref{figRecoveredImage} shows some of the reconstructed images ($k=100$) in the testing datasets. 
Peak Signal-to-Noise Ratios (PSNR) for reconstructed images are also provided. 
From Fig. \ref{figRecoveredImage}, one can see that the RPCA-DSWL algorithm consistently outperforms other algorithms, accurately describing the main characteristics of images and exhibiting superior robustness.
PSNRs of recovered images by the proposed RPCA-DSWL algorithm are the highest among all recovered images.
Besides that, visually the recovered images by the proposed algorithm are more identical to the original image than other recovered images.

\section{Conclusion}
Outliers pose a detrimental influence on the estimation of the projection matrix in PCA algorithms.
This paper proposes an algorithm termed RPCA-DSWL algorithm to mitigate outliers by learning sample weights that differentiate outliers from normal samples.
The proposed algorithm iteratively updates the sample weights, the sample mean, and the projection matrix. 
In learning sample weights, the projection variance, the reconstruction error and the distance to the mean are all taken into consideration. 
It is noteworthy that the weights assigned to outliers are smaller than those assigned to normal samples, demonstrating the capability of distinguishing outliers from normal samples.
The experimental simulation results demonstrated that the proposed algorithm has superb performance on sample reconstruction and classification tasks.

\bibliographystyle{IEEEtran.bst}
\bibliography{mybibliography.bib}

\begin{thebibliography}{10}
\providecommand{\url}[1]{#1}
\csname url@samestyle\endcsname
\providecommand{\newblock}{\relax}
\providecommand{\bibinfo}[2]{#2}
\providecommand{\BIBentrySTDinterwordspacing}{\spaceskip=0pt\relax}
\providecommand{\BIBentryALTinterwordstretchfactor}{4}
\providecommand{\BIBentryALTinterwordspacing}{\spaceskip=\fontdimen2\font plus
\BIBentryALTinterwordstretchfactor\fontdimen3\font minus \fontdimen4\font\relax}
\providecommand{\BIBforeignlanguage}[2]{{%
\expandafter\ifx\csname l@#1\endcsname\relax
\typeout{** WARNING: IEEEtran.bst: No hyphenation pattern has been}%
\typeout{** loaded for the language `#1'. Using the pattern for}%
\typeout{** the default language instead.}%
\else
\language=\csname l@#1\endcsname
\fi
#2}}
\providecommand{\BIBdecl}{\relax}
\BIBdecl

\bibitem{StandardPCA}
S.~Wold, K.~Esbensen, and P.~Geladi, ``{Principal component analysis},'' \emph{Chemometr. Intell. Lab. Syst.}, vol.~2, no.~1, pp. 37--52, 1987.

\bibitem{face}
M.~Turk and A.~Pentland, ``Face recognition using eigenfaces,'' in \emph{Proc. IEEE Conf. Comput. Vis. Pattern Recognit.}, 1991, pp. 586--591.

\bibitem{CV}
F.~De~La~Torre and M.~J. Black, ``A framework for robust subspace learning,'' \emph{Int. J. Comput. Vis.}, vol.~54, no.~1, pp. 117--142, 2003.

\bibitem{RPCA}
J.~Wright, Y.~Peng, Y.~Ma, A.~Ganesh, and S.~Rao, ``Robust principal component analysis: Exact recovery of corrupted low-rank matrices by convex optimization,'' in \emph{Proc. Int. Conf. Neural Inf. Process. Syst.}, Red Hook, NY, USA, 2009, pp. 2080--2088.

\bibitem{CRPCA}
Z.~Lin, M.~Chen, and Y.~Ma, ``The augmented lagrange multiplier method for exact recovery of corrupted low-rank matrices,'' \emph{arXiv preprint arXiv:1009.5055}, 2010.

\bibitem{SPCP}
Z.~Zhou, X.~Li, J.~Wright, E.~Candes, and Y.~Ma, ``Stable principal component pursuit,'' in \emph{Proc. IEEE Int. Symp. Inf. Theory}.\hskip 1em plus 0.5em minus 0.4em\relax IEEE, 2010, pp. 1518--1522.

\bibitem{RRPCP}
P.~P. Brahma, Y.~She, S.~Li, J.~Li, and D.~Wu, ``Reinforced robust principal component pursuit,'' \emph{IEEE Trans. Neural Netw. Learn. Syst.}, vol.~29, no.~5, pp. 1525--1538, 2017.

\bibitem{l1PCA}
Q.~Ke and T.~Kanade, ``{Robust $L_1$ norm factorization in the presence of outliers and missing data by alternative convex programming},'' in \emph{Proc. IEEE Comput. Soc. Conf. Comput. Vis. Pattern Recognit.}, vol.~1, 2005, pp. 739--746.

\bibitem{PCAL1}
N.~Kwak, ``{Principal component analysis based on $L_1$-norm maximization},'' \emph{IEEE Trans. Pattern Anal. Mach. Intell.}, vol.~30, no.~9, pp. 1672--1680, 2008.

\bibitem{PCALP}
------, ``Principal component analysis by $l_p$-norm maximization,'' \emph{IEEE Trans. Cybern.}, vol.~44, no.~5, pp. 594--609, 2013.

\bibitem{rotation}
A.~Y. Ng, ``Feature selection, $l_1$ vs. $l_2$ regularization, and rotational invariance,'' in \emph{Proc. Int. Conf. Mach. Learn.}, New York, NY, USA, 2004, pp. 78--85.

\bibitem{R1PCA}
C.~Ding, D.~Zhou, X.~He, and H.~Zha, ``{ $R_1$-PCA: rotational invariant $L_1$-norm principal component analysis for robust subspace factorization},'' in \emph{Proc. Int. Conf. Mach. Learn.}, 2006, pp. 281--288.

\bibitem{PCAOM}
F.~Nie, J.~Yuan, and H.~Huang, ``Optimal mean robust principal component analysis,'' in \emph{Proc. Int. Conf. Mach. Learn.}, vol.~32, no.~2, Bejing, China, 22--24 Jun 2014, pp. 1062--1070.

\bibitem{nongreedyl21}
F.~Nie, L.~Tian, H.~Huang, and C.~Ding, ``{Non-greedy $L_{2,1}$-norm maximization for principal component analysis},'' \emph{IEEE Trans. Image Process.}, vol.~30, pp. 5277--5286, 2021.

\bibitem{nogreedyl212}
F.~Nie, Z.~Wang, R.~Wang, Z.~Wang, and X.~Li, ``{Towards robust discriminative projections learning via non-greedy $L_{2,1}$-norm minmax},'' \emph{IEEE Trans. Pattern Anal. Mach. Intell.}, vol.~43, no.~6, pp. 2086--2100, 2021.

\bibitem{L2PPCA}
Q.~Wang, Q.~Gao, X.~Gao, and F.~Nie, ``{$L_{2,p}$-norm based PCA for image recognition},'' \emph{IEEE Trans. Image Process.}, vol.~27, no.~3, pp. 1336--1346, 2018.

\bibitem{2attentivePCA}
D.~Wu, H.~Zhang, F.~Nie, R.~Wang, C.~Yang, X.~Jia, and X.~Li, ``Double-attentive principle component analysis,'' \emph{IEEE Signal Process. Lett.}, vol.~27, pp. 1814--1818, 2020.

\bibitem{adaptiveLossPCA}
J.~Bian, D.~Zhao, F.~Nie, R.~Wang, and X.~Li, ``Robust and sparse principal component analysis with adaptive loss minimization for feature selection,'' \emph{IEEE Trans. Neural Netw. Learn. Syst.}, pp. 1--14, 2022, early access.

\bibitem{anglePCA}
Q.~Wang, Q.~Gao, X.~Gao, and F.~Nie, ``Angle principal component analysis,'' in \emph{Proc. Int. Joint Conf. Artif. Intell.}, 2017, pp. 2936--2942.

\bibitem{PCADI}
Y.~Gao, T.~Lin, Y.~Zhang, S.~Luo, and F.~Nie, ``Robust principal component analysis based on discriminant information,'' \emph{IEEE Trans. Knowl. Data Eng.}, vol.~35, no.~2, pp. 1991--2003, 2023.

\bibitem{WPCA}
Y.~Koren and L.~Carmel, ``Robust linear dimensionality reduction,'' \emph{IEEE Trans. Vis. Comput. Graph.}, vol.~10, no.~4, pp. 459--470, 2004.

\bibitem{PCAAN}
R.~Zhang and H.~Tong, ``Robust principal component analysis with adaptive neighbors,'' in \emph{Proc. Adv. Neural Inf. Process. Syst.}, vol.~32.\hskip 1em plus 0.5em minus 0.4em\relax Curran Associates, Inc., 2019.

\bibitem{TRPCA}
F.~Nie, D.~Wu, R.~Wang, and X.~Li, ``Truncated robust principle component analysis with a general optimization framework,'' \emph{IEEE Trans. Pattern Anal. Mach. Intell.}, vol.~44, no.~2, pp. 1081--1097, 2020.

\bibitem{DRPCA}
F.~Nie, S.~Wang, Z.~Wang, R.~Wang, and X.~Li, ``Discrete robust principal component analysis via binary weights self-learning,'' \emph{IEEE Trans. Neural Netw. Learn. Syst.}, pp. 1--14, 2022, early access.

\bibitem{pcajrp}
S.~Wang, F.~Nie, Z.~Wang, R.~Wang, and X.~Li, ``Robust principal component analysis via joint reconstruction and projection,'' \emph{IEEE Trans. Neural Netw. Learn. Syst.}, vol.~35, no.~5, pp. 7175--7189, 2024.

\bibitem{nocedal1999numerical}
J.~Nocedal and S.~J. Wright, \emph{Numerical optimization}.\hskip 1em plus 0.5em minus 0.4em\relax Springer, 1999.

\bibitem{GENER}
J.~Liu, S.~Chen, Z.-H. Zhou, and X.~Tan, ``Generalized low-rank approximations of matrices revisited,'' \emph{IEEE Trans. Neural Netw.}, vol.~21, no.~4, pp. 621--632, 2010.

\bibitem{YALE}
A.~S. Georghiades, P.~N. Belhumeur, and D.~J. Kriegman, ``From few to many: Illumination cone models for face recognition under variable lighting and pose,'' \emph{IEEE Trans. Pattern Anal. Mach. Intell.}, vol.~23, no.~6, pp. 643--660, 2001.

\bibitem{orl}
M.~Turk and A.~Pentland, ``Eigenfaces for recognition,'' \emph{Journal of Cognitive Neuroscience}, vol.~3, no.~1, pp. 71--86, 1991.

\bibitem{umist}
H.~Wechsler, J.~P. Phillips, V.~Bruce, F.~F. Soulie, and T.~S. Huang, \emph{Face Recognition: From Theory to Applications}.\hskip 1em plus 0.5em minus 0.4em\relax Springer Science \& Business Media, 2012, vol. 163.

\end{thebibliography}
\section*{APPENDIX A}

In this appendix, we solve the following optimization problem
\begin{align}
\notag
&\max_{a_1,a_2,\dots,a_n} \frac{1}{n}\sum_{i=1}^{n} a_i \|\boldsymbol{P}^T(\boldsymbol{x}_i-\boldsymbol{m})\|_2^2 + \tau_a\left(-\sum_{i=1}^n a_i \textup{ln}a_i\right)\\
\notag
&s.t. \quad \sum_{i=1}^{n} a_i=1,\\
&~~\qquad0\leq a_i\leq 1,\quad i=1,2,\cdots, n.
\end{align}

The standard form of this optimization problem is
\begin{align}
\notag
&\min_{a_1,a_2,\dots,a_n} -\frac{1}{n}\sum_{i=1}^{n}a_i \|\boldsymbol{P}^T(\boldsymbol{x}_i-\boldsymbol{m})\|_2^2 + \tau_a\left(\sum_{i=1}^n a_i \textup{ln}a_i\right)\\
\notag
&s.t. \quad \sum_{i=1}^{n} a_i=1,\\
&~~\qquad0\leq a_i\leq 1, \quad  i=1,2,\cdots, n.
\end{align}

The Lagrangian function is
\begin{align}
\notag
    L=&-\frac{1}{n}\sum_{i=1}^{n} a_i \|\boldsymbol{P}^T(\boldsymbol{x}_i-\boldsymbol{m})\|_2^2 + \tau_a\left(\sum_{i=1}^n a_i \textup{ln}a_i\right)\\
&+\nu\left(\sum_{i=1}^{n} a_i-1\right)+\sum_{i=1}^n \lambda_i(a_i- 1)+\sum_{i=1}^n\eta_i(-a_i),
\end{align}
where $\nu$, $\lambda_i\geq 0$, $\eta_i\geq 0$ are Lagrangian dual variables. According to the Karush–Kuhn–Tucker conditions(KKT) conditions, there are
\begin{align}
    &\sum_{i=1}^na_i=1,\\
    &\lambda_i\geq 0, \eta_i\geq0, \quad i=1,2\cdots, n,\\
    \label{CSC1}
    &\lambda_i(a_i-1)=0, \quad i=1,2\cdots, n,\\
    \label{CSC2}
    &\eta_i a_i=0,\quad i=1,2\cdots, n,\\
    \notag
    &\frac{\partial L}{\partial a_i} =  -\frac{1}{n}\|\boldsymbol{P}^T(\boldsymbol{x}_i-\boldsymbol{m})\|_2^2 + \tau_a\left(\textup{ln}a_i+1\right)\\
    \label{deltaL}
&\qquad\quad +\nu+\lambda_i-\eta_i=0,\quad i=1,2,\cdots, n,
\end{align}
where (\ref{CSC1}) and (\ref{CSC2}) are complementary slackness conditions.

From (\ref{deltaL}), one can get
\begin{align}
    a_i = \textup{exp}\left(\frac{1}{n\tau_a} \|\boldsymbol{P}^T(\boldsymbol{x}_i-\boldsymbol{m})\|_2^2 -1
-\frac{\nu+\lambda_i-\eta_i}{\tau_a}\right),
\end{align}
from which one can see that $0<a_i<1$ because $a_i$ is in an exponential form. Taking together $0<a_i<1$, $\lambda_i(a_i-1)=0$ in (\ref{CSC1}) and $\eta_i a_i=0$ in (\ref{CSC2}), there are $\lambda_i=0$ and $\eta_i=0$ for $i=1,2\cdots, n$. Together with the constraint $\sum_{i=1}^na_i=1$, one gets
\begin{align}
    a_i = \frac{\textup{exp}\left(\frac{1}{n\tau_a} \|\boldsymbol{P}^T(\boldsymbol{x}_i-\boldsymbol{m})\|_2^2\right)}{\sum_{j=1}^n\textup{exp}\left(\frac{1}{n\tau_a} \|\boldsymbol{P}^T(\boldsymbol{x}_j-\boldsymbol{m})\|_2^2\right)}.
\end{align}

\end{document}